\documentclass{article}

\usepackage{microtype}
\usepackage{graphicx}
\usepackage{subcaption}
\usepackage{booktabs} 
\usepackage{fontawesome5} 
\usepackage{hyperref}
\usepackage{todonotes}

\usepackage[preprint]{icml2026}

\usepackage{amsmath}
\usepackage{amssymb}
\usepackage{mathtools}
\usepackage{amsthm}

\usepackage[capitalize,noabbrev]{cleveref}

\theoremstyle{plain}

\theoremstyle{definition}

\theoremstyle{remark}

\usepackage[table]{xcolor}
\usepackage{url}
\usepackage{booktabs}
\usepackage{graphicx} 
\usepackage{caption}
\usepackage{xcolor}
\definecolor{red}{RGB}{255, 0, 0}
\definecolor{blue}{RGB}{0, 0, 255}
\definecolor{magenta}{RGB}{255, 0, 255}
\definecolor{good}{RGB}{54, 183, 0}
\definecolor{bad}{RGB}{157, 44, 0}
\definecolor{comment}{RGB}{159, 43, 104}
\usepackage{xspace}
\usepackage{colortbl}
\usepackage{subcaption}
\usepackage{multirow} 
\usepackage{tikz}
\usepackage{makecell}
\usepackage{algorithm}
\usepackage{listings}
\usepackage{wrapfig}
\usepackage{titlesec}
\usepackage{enumitem}
\usepackage{pifont}
\usepackage[breakable]{tcolorbox}
\titlespacing\section{0pt}{0pt}{1pt}
\titlespacing\subsection{0pt}{0pt}{0pt}
\titlespacing\subsubsection{0pt}{0pt}{0pt}
\titlespacing\paragraph{0pt}{0pt}{4pt}

\newcommand{\llmname}[1]{{\fontfamily{pcr}\selectfont {#1}}\xspace}

\hypersetup{
    colorlinks=true,
    linkcolor=red,
    citecolor=cyan,
    filecolor=magenta,      
    urlcolor=magenta,
    }

\newcommand{\algname}{\textsc{CircuitProbe}\xspace}
\newcommand{\tabref}[1]{Table~\ref{#1}}
\newcommand{\figref}[1]{Fig.~\ref{#1}}
\newcommand{\eqnref}[1]{\text{Eq.}~(\ref{#1})}
\newcommand{\secref}[1]{\S\ref{#1}}
\newcommand{\appref}[1]{Appendix~\ref{#1}}

\newcommand{\Iv}{\mathcal{I}_V}
\newcommand{\Iq}{\mathcal{I}_Q}

\usepackage{pgf}

\newtcolorbox{takeawaybox}[1][]{
    breakable, 
    colback=black!5, 
    colframe=black!70, 
    boxrule=1pt, 
    arc=2mm, 
    fonttitle=\bfseries, 
    #1 
}

\newcommand{\perf}[2][\red]{%
  \pgfmathsetmacro{\accuracy}{100 - #2}%
  \pgfmathprintnumber[fixed, fixed zerofill, precision=1]{\accuracy}%
  #1{#2}%
}
\newcommand{\perfmax}[2][\blue]{%
  \pgfmathsetmacro{\accuracy}{100 - #2}%
  \pgfmathprintnumber[fixed, fixed zerofill, precision=1]{\accuracy}%
  #1{#2}%
}
\newcommand{\val}[2]{%
  \ifdim #1pt < 100.0pt
    #1$_{\textcolor{red}{\,\downarrow #2}}$%
  \else
    \textbf{#1}%
  \fi
}

\newcommand{\gain}[2]{%
  #1$_{\textcolor{blue}{\,\uparrow #2}}$%
}
\newcommand{\loss}[2]{%
  #1$_{\textcolor{red}{\,\downarrow #2}}$%
}
\icmltitlerunning{\algname: Tracing Visual Temporal Evidence Flow in Video Language Models}
\begin{document}
\twocolumn[
  \icmltitle{\algname: Tracing Visual Temporal Evidence Flow in \\ Video Language Models}
  \vspace{-0.2in}

  \icmlsetsymbol{equal}{*}
  \begin{icmlauthorlist}
    \icmlauthor{Yiming Zhang~\dag}{equal,hfips,ustc}
    \icmlauthor{Zhuokai Zhao}{equal,uc}
    \icmlauthor{Chengzhang Yu}{scut}
    \icmlauthor{Kun Wang}{ntu}
    \icmlauthor{Zhendong Chu}{sqa}
    \icmlauthor{Qiankun Li}{ntu}
    \icmlauthor{Zihan Chen}{hfips,ustc}
    \icmlauthor{Yang Liu}{ntu}
    \icmlauthor{Zenghui Ding}{hfips}
    \icmlauthor{Yining Sun}{hfips}
    \icmlauthor{Qingsong Wen}{sqa}
  \end{icmlauthorlist}
      \begin{center}
          \faGithub~\textbf{Project:} \href{https://jam1ezhang.github.io/CircuitProbe/}{circuitprobe.github.io}
      \end{center}
    \vspace{-0.1in}

    \icmlaffiliation{hfips}{HFIPS, Chinese Academy of Sciences}
    \icmlaffiliation{ustc}{University of Science and Technology of China}
    \icmlaffiliation{uc}{University of Chicago}
    \icmlaffiliation{scut}{South China University of Technology}
    \icmlaffiliation{sqa}{Squirrel AI}
    \icmlaffiliation{ntu}{Nanyang Technological University}

  \icmlcorrespondingauthor{Zenghui Ding}{dingzenghui@iim.ac.cn}
  \icmlcorrespondingauthor{Kun Wang}{wk520529@mail.ustc.edu.cn}
  \icmlkeywords{Machine Learning, ICML}
  \vskip 0.1in

]
\printAffiliationsAndNotice{\dag Work done at Squirrel AI}  
\begin{abstract}
Autoregressive large vision--language models (LVLMs) interface video and language by projecting video features into the LLM's embedding space as continuous visual token embeddings.
However, it remains unclear \textit{where} temporal evidence is represented and \textit{how} it causally influences decoding.
To address this gap, we present \textbf{\algname}, a circuit-level analysis framework that dissects the end-to-end video-language pathway through two stages:
\textit{(i) Visual Auditing}, which localizes object semantics within the projected video-token sequence and reveals their causal necessity via targeted ablations and controlled substitutions;
and \textit{(ii) Semantic Tracing}, which uses logit-lens probing to track the layer-wise emergence of object and temporal concepts, augmented with temporal frame interventions to assess sensitivity to temporal structure.
Based on the resulting analysis, we design a targeted surgical intervention that strictly follows our observations: identifying temporally specialized attention heads and selectively amplifying them within the critical layer interval revealed by Semantic Tracing.
This analysis-driven intervention yields consistent improvements (up to 2.4\% absolute) on the temporal-heavy TempCompass benchmark, validating the correctness, effectiveness, and practical value of the proposed circuit-level analysis for temporal understanding in LVLMs. 
%
%
%
%
\end{abstract}
\vspace{-0.1in}
\vspace{-0.1in}
\section{Introduction}\label{sec:introduction}
Large vision--language models (LVLMs) have become a dominant paradigm for multimodal understanding by integrating visual inputs (images or videos) with natural language to produce free-form textual outputs~\citep{liu2023llava, liu2023improvedllava, maazvideochatgptdetailedvideo2023, Zhang2024BeyondTD}.
Most LVLMs follow a modular design that couples a pre-trained visual encoder with a pre-trained large language model (LLM) through a trainable adapter, or projector~\citep{liBLIPBootstrappingLanguageImage2022, moonAnyMALEfficientScalable2023, chenminigptv2largelanguage2023}.
This projector converts visual features into continuous token embeddings (often referred to as soft prompts) that are compatible with the LLM's input space, enabling cross-modal alignment and downstream reasoning~\citep{merullo2023linearlymappingimagetext, liu2024llavanext, yu2025cafe, pan2025transfer}.
\vspace{-0.05in}

Understanding \emph{how} video evidence influences language decoding is increasingly important for two reasons.
First, it clarifies what information is actually transferred through the visual--language interface, which directly affects vision--language alignment quality~\citep{Radford2021LearningTV, Zhang2024RankCLIPRL}.
Second, it provides actionable guidance for designing LVLMs with robust reasoning behaviors beyond benchmark-driven improvements~\citep{pangFrozenTransformersLanguage2024, wooDontMissForest2024, parkBridgingVisionLanguage2024, wang2025comprehensive}.
\vspace{-0.05in}

While recent interpretability work has substantially advanced our understanding of image-based LVLM representations and visual-text integration\citep{neoInterpretingVisualInformation2024}, it remains less clear how these tools extend to \textit{video} inputs.
Unlike images, video understanding requires the model to aggregate evidence across frames and integrate cues that are distributed over time.
The challenge is that video-derived soft prompts do not correspond directly to discrete language tokens, making it difficult to attribute model outputs to specific temporal evidence or to isolate where temporal reasoning succeeds or fails~\citep{merullo2023linearlymappingimagetext, yinsurveymultimodallarge2023}.
\vspace{-0.05in}

To address this gap, in this paper, we introduce \textbf{\algname}, a circuit-based tracing framework that operationalizes two complementary analytic probes.
First, \textit{visual auditing} localizes task-critical semantics within the projected video-token sequence and tests their causal necessity through targeted token ablations and controlled substitutions.
%
Second, \textit{semantic tracing} tracks the layer-wise emergence of language-aligned object and action concepts by logit-lens–style~\citep{InterpretingGPTLogit} probe and shows how the temporal structure affects semantic consolidation.
Motivated by these observations, we design targeted interventions that follow our findings to test whether the identified structures are causally responsible for temporal reasoning.
Our contributions can be summarized as follows:
\begin{itemize}[leftmargin=*, noitemsep, topsep=0pt]
    \item We show that video semantics remain strongly localized to object-aligned token positions after projection: ablating object tokens causes substantially larger performance degradation than removing matched control tokens.
    \item We find that object and action concepts become reliably language-aligned primarily in mid-to-late layers of the LLM, defining a critical interval where video-conditioned semantics are consolidated. 
    We further observe behaviors consistent with frame-wise state aggregation rather than explicit temporal order modeling, motivating targeted temporal interventions.
    \item Using attention-based temporal metrics, we isolate temporally specialized attention heads and show that selectively amplifying them within the identified critical interval yields consistent improvements on temporal-heavy benchmark such as TempCompass~\citep{liu2024tempcompass}. 
    These results validate the correctness, effectiveness, and practical value of our circuit-level analysis.
    %
\end{itemize}
\begin{figure*}[t]
    \centering
    \includegraphics[width=\linewidth]{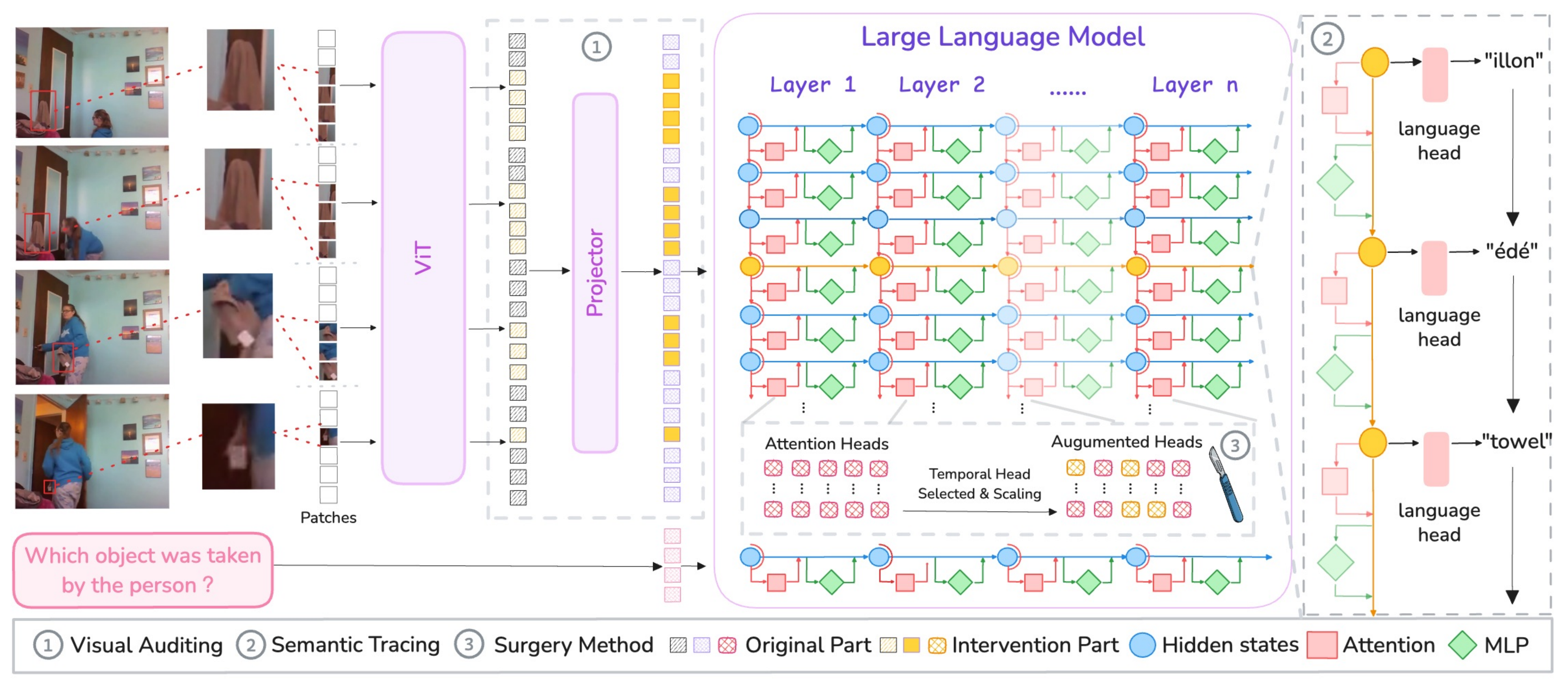}
    \vspace{-0.2in}
    \caption{
        \algname consists of two circuits: 
        \ding{192} \textit{Visual Auditing}, which localizes task-critical object semantics within the projected video-token sequence, 
        and \ding{193} \textit{Semantic Tracing}, which tracks the layer-wise emergence of language-aligned concepts. 
        Based on these observations, Surgical Recovery amplifies \textit{temporal attention heads} within the identified layers, yielding gains on temporal benchmarks.
        %
    }
    \label{fig:circuit_probe}
    \vspace{-0.25in}
\end{figure*}
%


\section{Related work} 
\paragraph{Interpretability of LLM.} 
Recent research on LLM interpretability has largely focused on text-only models~\citep{Singh2024RethinkingII,wang2025large}. Mainstream approaches include circuit analysis, which simplifies models into sparse circuits to understand mechanisms~\citep{hanna2023does,yao2024knowledge}, and causal tracing, which tracks information flow to analyze module contributions and edit factual knowledge\citep{yu2024interpreting,sharma2024locating}, as pioneered by ROME~\citep{meng2022locating}. Another prominent technique is unembedding space projection, which probes internal states by projecting them into the token space~\citep{geva2021transformer,dar2023analyzing}. %
While these methods have enhanced our understanding of language models, they focus predominantly on textual processing. Multimodal interpretability, particularly the analysis of visual-textual integration, remains an under-explored frontier.
\paragraph{Interpretability of LVLMs.} 
Existing work on visual model interpretability typically focuses on two areas. 
The first examines visual encoder embedding generation, including how frozen transformers process visual tokens~\citep{neoInterpretingVisualInformation2024,kaduri2024whatsimagedeepdivevision}, how misalignment causes hallucinations~\citep{jiang2025interpretingeditingvisionlanguagerepresentations}, and how modality spaces are bridged~\citep{pangFrozenTransformersLanguage2024, wooDontMissForest2024, parkBridgingVisionLanguage2024}. 
The second investigates visual-textual interactions in multimodal systems, often by identifying shortcomings on static images~\citep{tongEyesWideShut2024,Verma2024CrossModalPI}, extracting visual concept subgraphs~\citep{rajaram2024automaticdiscoveryvisualcircuits}, or reverse-engineering ViTs~\citep{vilas2023analyzingvisiontransformersimage}. 
Recent work has begun to address video data, identifying temporal reasoning bottlenecks~\citep{li2024temporal} and developing new interpretability tools~\citep{joseph2025prisma}. 
However, these studies do not examine how visual temporal evidence is consolidated over time within the LLM, particularly in the context of video input. This leaves a gap in understanding how temporal relationships are processed in LVLMs, which our work aims to fill by investigating the role of temporal attention heads and their impact on model performance.
%
\section{\algname}\label{sec:method}
\vspace{-0.05in}
\subsection{Preliminary Setup}\label{subsec:prelim_setup}
Before introducing the two analytic circuits of \algname, we first establish a unified notation, dataset construction protocol, and attention-based primitives that will be shared across all subsequent analyses.  
We evaluate on three representative LVLM backbones covering both image- and video-trained 7B variants: 
the \llmname{LLaVA-NeXT} family~\citep{li2024llava}, 
the \llmname{LLaVA-OneVision} family~\citep{li2024llavaonevisioneasyvisualtask}, 
and \llmname{Qwen2.5-VL}~\citep{bai2025qwen2}. 
For brevity, we denote \textit{image}-trained and \textit{video}-trained model variants using the suffixes \llmname{\textit{-I}} and \llmname{\textit{-V}}, respectively.
%
%

\paragraph{Model input representation.}
Given a video sampled into $T$ frames and a textual query $Q$, we construct the combined input embedding sequence 
$\mathbf{X}_0 = [\mathbf{X}_{\mathrm{sys}}; \mathbf{X}_V; \mathbf{X}_Q] \in \mathbb{R}^{N \times d}$, 
where $N$ is the total number of tokens and $d$ is the hidden dimension of the LLM.
Here, $\mathbf{X}_{\mathrm{sys}}$ denotes system tokens, $\mathbf{X}_Q$ corresponds to the query tokens, and $\mathbf{X}_V$ represents the video-conditioned tokens injected into the LLM.
Specifically, video tokens $\mathbf{X}_V \in \mathbb{R}^{N_V \times d}$ are obtained by encoding each frame into $P$ visual patches using the vision encoder $f_{\mathrm{C}}$ followed by the trainable adapter (projector).
This results in $N_V=T \times P$ video tokens, which we index by 
$\mathcal{I}_V = \bigcup_{t=1}^T \mathcal{I}_{V,t}$,
where $\mathcal{I}_{V,t}$ denotes the subset of tokens originating from frame $t$. 
We denote all non-video token indices (system and query tokens) by $\mathcal{I}_Q$. 
%
%

%

\paragraph{Circuit-based analysis paradigm.}
With this setup in place, we adopt a circuit-based analysis paradigm to dissect the internal computation of LVLMs.
Our approach explicitly manipulates subsets of video tokens and traces their transformation across layers, enabling us to test concrete predictions about information localization and semantic emergence.
%
Importantly, both visual auditing (Circuit \ding{192}, \secref{subsec:visual_auditing}) and semantic tracing (Circuit \ding{193}, \secref{subsec:semantic_tracing}), operate on this shared foundation, ensuring that conclusions drawn from different probes are directly comparable.
Unless otherwise specified, our analyses focus on the prefill stage of autoregressive decoding and on video-token positions, where visual information is first injected into the LLM and can directly influence downstream reasoning.
\paragraph{Dataset curation and probing setup.}
To probe genuine temporal understanding, we curate a subset of the STAR benchmark~\citep{wuSTARBenchmarkSituated} focusing on two tasks: \textit{Interaction} (action-object binding) and \textit{Sequence} (chronological dependencies). To strictly isolate visual reasoning from language priors, we enforce three-step filtering process: (1) \textbf{Indispensability}: We retain only samples where the answer requires integrating video information, filtering out those solvable by single frames or common sense. (2) \textbf{Visual Clarity}: We select key frames that clearly depict the target actions or objects. (3) \textbf{Counterfactual Validity}: We verify that the model answers correctly on the original input but fails when the critical object is masked. This process yields a controlled testbed for analyzing causal visual integration over time. We employ two complementary probing formats:
\begin{enumerate}[leftmargin=*, noitemsep, topsep=0pt]
    \vspace{-0.05in}
    \item \textit{Open-ended questions.}
    (e.g., "Which object...") with prefix-forcing to trace specific semantic concepts;
    \item \textit{Close-ended questions.} 
    ("Yes/No") to measure reasoning robustness via prediction flips under intervention. 
    \vspace{-0.05in}
    %
\end{enumerate}
Unless otherwise specified, Circuit~\ding{192} reports results under both formats, while Circuit~\ding{193} focuses on the open-ended format only for concept-level tracing. Representative example frames and more dataset details are shown in \appref{app:Analysis Setup}.

\begin{table*}[htbp]
\centering
\caption{
Accuracy drop ($\downarrow$) from visual token ablation.
All performance drops are \textit{relative} to the \textit{No Ablation Baseline} (which has a drop of 0).
For object-based ablation, the under-script $_{bf}$ represents the number of spatially adjacent tokens, as described in object tokens with buffer.
We see that ablating object tokens leads to significantly higher drops compared to random tokens, verifying semantic localization.
}
\vspace{-0.05in}
\label{tab:ablation_improved}
\resizebox{\textwidth}{!}{
\begin{tabular}{@{}lc cc cc cc cc@{}}
\toprule
& & \multicolumn{2}{c}{\textbf{LLaVA-NeXT-I}} & \multicolumn{2}{c}{\textbf{LLaVA-NeXT-V}} & \multicolumn{2}{c}{\textbf{LLaVA-OV-I}} & \multicolumn{2}{c}{\textbf{LLaVA-OV-V}} \\
\cmidrule(lr){3-4} \cmidrule(lr){5-6} \cmidrule(lr){7-8} \cmidrule(lr){9-10}
\textbf{Type} & \textbf{Avg Tokens} & Open & Close & Open & Close & Open & Close & Open & Close \\
\midrule
No Ablation& 0 & \perf{0} & \perf{0} & \perf{0} & \perf{0} & \perf{0} & \perf{0} & \perf{0} & \perf{0} \\
\midrule
Register & 13 & \perf{3.4} & \perf{1.7} & \perf{2.0} & \perf{1.8} & \perf{15.3} & \perf{1.8} & \perf{9.7} & \perf{3.1} \\
\midrule
\multirow{3}{*}{Random} 
 & 100 & \perf{2.3} & \perf{0.9} & \perf{2.3} & \perf{1.6} & \perf{14.8} & \perf{2.0} & \perf{9.5} & \perf{2.9} \\
 & 500 & \perf{6.0} & \perf{1.4} & \perf{6.7} & \perf{1.3} & \perf{15.1} & \perf{2.3} & \perf{10.6} & \perf{3.5} \\
 & 900 & \perf{10.6} & \perf{2.0} & \perf{10.7} & \perf{1.6} & \perf{14.7} & \perf{2.8} & \perf{9.6} & \perf{2.9} \\
\midrule
Object & 354 & \perf{81.9} & \perf{54.1} & \perf{82.7} & \perf{37.7} & \perf{59.5} & \perf{32.1} & \perf{39.2} & \perf{21.2} \\
Object$_{(bf=1)}$ & 476 & \perf{90.0} & \perf{65.5} & \perf{87.5} & \perf{50.7} & \perf{66.4} & \perf{46.7} & \perf{47.4} & \perf{32.9} \\
Object$_{(bf=2)}$ & 634 & \perfmax{90.7} & \perfmax{72.2} & \perfmax{92.6} & \perfmax{55.1} & \perfmax{69.1} & \perfmax{54.1} & \perfmax{48.3} & \perfmax{37.0} \\
\bottomrule
\end{tabular}
}
\vspace{-0.1in}
\end{table*}
\subsection{Circuit \ding{192}: Visual Auditing}\label{subsec:visual_auditing}
As illustrated in \figref{fig:circuit_probe}, Circuit~\ding{192} seeks to identify where task-critical visual semantics reside after projection into the LLM input space, and what form those semantics take for downstream reasoning.
We address these questions by using controlled manipulations of projected video tokens to identify where task-critical visual semantics reside in the input sequence and how these semantics are represented for downstream language decoding.
\vspace{-0.05in}
\subsubsection{Where does task-critical visual information 
reside?}\label{subsec:observation_1}
\vspace{-0.05in}
We begin by asking whether object-level visual semantics are diffusely encoded across the projected video-token sequence $X_V$.
Inspired by \citep{neoInterpretingVisualInformation2024}, Following the notation introduced in \secref{subsec:prelim_setup}, let $\mathcal{I}_V$ denote the index set of all projected video tokens in $X_V$. 
We define an object-centric index set $\mathcal{I}_O \subset \mathcal{I}_V$ consisting of tokens whose corresponding image patches are fully contained within the ground-truth bounding box of the target object. 
The remaining video tokens $\mathcal{I}_V \setminus \mathcal{I}_O$ serve as a pool from which control subsets are constructed.
\paragraph{Intervention.}
To causally test which subsets of video tokens carry task-critical semantics, we remove information from selected token subsets by \textit{replacement rather than deletion}. 
Specifically, for any subset $\mathcal{S} \subset \mathcal{I}_V$, we replace the corresponding token embeddings in $X_V$ with an \textit{uninformative visual embedding} $x_{\text{null}}$, defined as the mean visual embedding computed over all visual tokens extracted from 10,000 images randomly sampled from the ImageNet~\citep{deng2009imagenet} validation set. 
This replacement preserves embedding scale and distribution, while selectively removing semantic content, thereby avoiding artifacts introduced by out-of-distribution perturbations.
All non-video tokens $X_{\text{sys}}$, $X_Q$, and all remaining video tokens, are left unchanged. 
We consider four ways of selecting the ablated token subset $\mathcal{S}$:
\begin{enumerate}[leftmargin=*, noitemsep, topsep=0pt, after=\vspace{-0.1\baselineskip}]
    \item \textit{Register Tokens.} Following~\citet{darcet2024visiontransformersneedregisters}, 
    within each key frame, we select tokens whose $L_2$ norms are more than two standard deviations above the mean token norm, and ablate these tokens as a control subset.
    \item \textit{Random Tokens.} As a baseline, we randomly sample a subset $\mathcal{I}_R \subset \mathcal{I}_V$ such that $|\mathcal{I}_R| = |\mathcal{I}_O|$. 
    This controls for the effect of removing an equal number of video tokens without targeting any specific semantic content.
    \item \textit{Object Tokens.} We ablate the object-centric token set $\mathcal{I}_O$, corresponding to image patches fully contained within the object’s bounding box.
    \item \textit{Object Tokens with Buffer.} To account for spatial imprecision and contextual spillover, we expand $\mathcal{I}_O$ to include spatially adjacent tokens. 
    We evaluate two buffer sizes: 1-Buffer, which includes immediately neighboring tokens, and 2-Buffer, which includes all tokens within a two-position radius of any object token.
\end{enumerate}
\paragraph{Observation.}
As shown in \tabref{tab:ablation_improved}, ablating object-aligned tokens induces a dramatically larger degradation in performance than ablating any of the control token subsets. 
This causal asymmetry persists even when the control set contains an equal or substantially larger number of tokens. 
For example, on \llmname{LLaVA-NeXT-V}~\citep{li2024llava}, replacing 573 object tokens in $\mathcal{I}_O$ leads to a 92.6\% drop in open-ended accuracy, whereas replacing 900 randomly selected tokens in $\mathcal{I}_R$ reduces performance by only 10.7\%. 
Similar trends are observed for register tokens and buffered object tokens, and the effect is consistent across both open-ended and close-ended evaluation protocols, as well as across \llmname{LLaVA-NeXT}~\citep{li2024llava} and \llmname{LLaVA-OneVision}~\citep{li2024llavaonevisioneasyvisualtask} variants.

\subsubsection{What representation does the LLM use for object semantics?}
As illustrated in \figref{fig:circuit_probe}, Circuit~\ding{192} operates on the projected video-token sequence $X_V$ before any cross-layer transformation occurs.
While \secref{subsec:observation_1} establishes where task-critical visual semantics are localized in the input sequence, we now ask a complementary question: in what representational form does the LLM most effectively use this information for downstream reasoning?
Specifically, we test whether object-level semantics must be conveyed through native visual token embeddings, or whether equivalent information can be supplied directly in the language embedding space.

\paragraph{Intervention.}
Let $\mathcal{I}_O \subset \mathcal{I}_V$ denote the object-aligned token indices identified in \secref{subsec:observation_1}. 
Instead of masking, we perform a controlled substitution by replacing the visual embeddings at indices $\mathcal{I}_O$ with the embedding of the object's textual label (e.g., replacing visual tokens corresponding to a towel with the embedding of the word ``towel'').
%
%
All other inputs, including token positions, ordering, system tokens $X_{\text{sys}}$, and query tokens $X_Q$ are left unchanged. 
We compare this condition against both the original visual-only input and the object-ablation baseline introduced in \secref{subsec:observation_1}.

\begin{figure}[ht]
    \centering
    \includegraphics[width=0.95\linewidth]{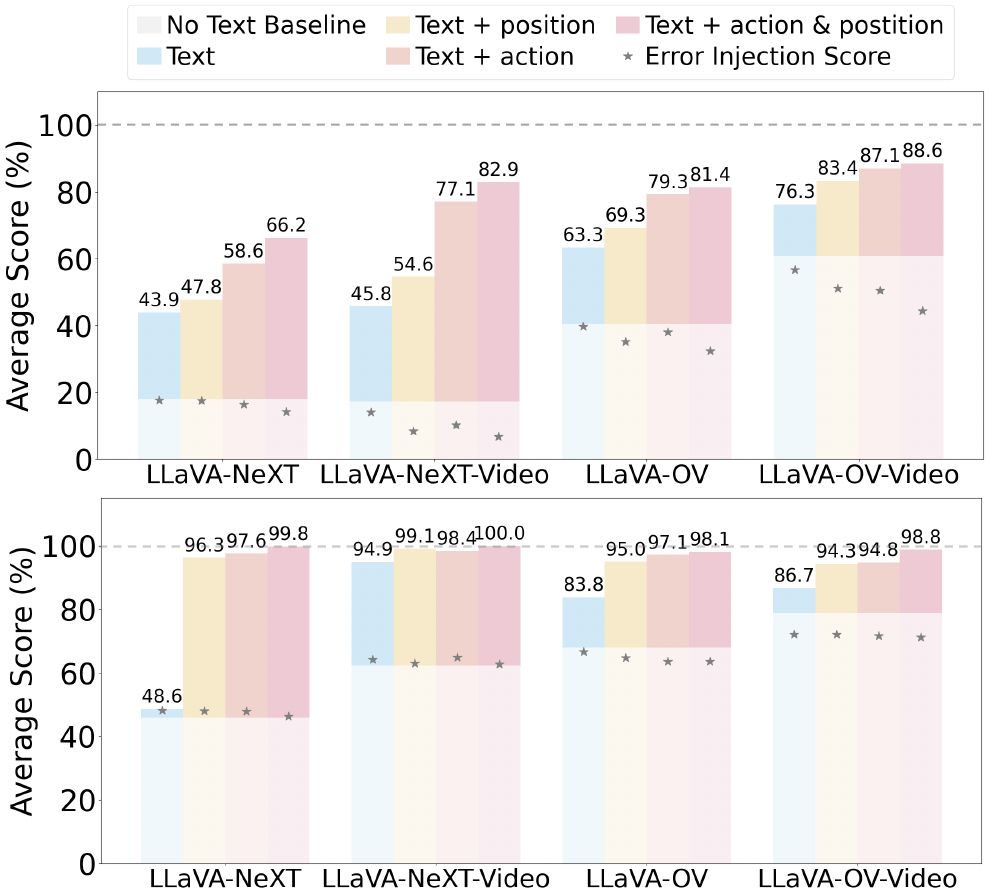}
    \caption{
        \textbf{\textit{Text injection experiment}}. 
        The result shows the performance change when visual object tokens are replaced by their corresponding embedded textual labels. Error injection indicates wrong concept injection. 
        %
        %
    }
    \label{fig:text_inject}
    \vspace{-0.2in}
\end{figure}

\paragraph{Observation.}
As shown in \figref{fig:circuit_probe}, replacing object-aligned visual tokens with their corresponding textual embeddings partially recovers the performance loss caused by object ablation. Moreover, substituting visual object tokens with their corresponding text embeddings only partially recovers performance, indicating that native multimodal evidence carries complementary cues beyond discrete label.
In the open-ended question setting, examples that collapse under object ablation (e.g., 17.3\% accuracy) recover to 82.9\% after text injection.
Conversely, \figref{fig:text_inject} shows that injecting an incorrect textual label consistently degrades performance, confirming that the gains arise from supplying the correct semantic content rather than from generic perturbation effects.
This behavior is consistent across LVLM variants and across both open-ended and close-ended evaluation formats.

\begin{tcolorbox}[
  colback=black!5,
  colframe=black!70,
  title=\textbf{Circuit~\ding{192} Takeaway},
  fonttitle=\bfseries,
  boxrule=0.6pt,
  arc=2mm,
  left=6pt,
  right=6pt,
  top=6pt,
  bottom=6pt,
  before skip=6pt,
  after skip=6pt
]
Object semantics are localized in a small set of object-aligned video tokens, but effective reasoning relies on their language-aligned conceptual representations.
\end{tcolorbox}

\subsection{Circuit~\ding{193}: Semantic Tracing }
\label{subsec:semantic_tracing}
Circuit~\ding{192} showed that task-critical visual information enters LLM through the compact set of object-aligned video tokens, and that effective reasoning ultimately relies on language-aligned conceptual representations. 
We builds on these constraints and asks the process-level question: 
\textit{How do object and temporal semantics carried by video tokens evolve across LLM layers before influencing decoding?}

Answering this question is essential for identifying where temporal reasoning can be causally intervened upon. 
If visual semantics become usable for language decoding only after a particular stage of processing, then only mechanisms operating at that stage should meaningfully affect temporal understanding---the prediction we later test via targeted surgery in \secref{sec:surgery_recovery}.

\subsubsection{Logit-lens probing: measuring concept evolving across layers}
To trace when video-conditioned semantic concepts become expressed in the language space, we employ the logit-lens method~\citep{InterpretingGPTLogit}. Specifically, we project the intermediate hidden states $h_i^{(\ell)}$ of visual tokens into the vocabulary space using the frozen output head (see \appref{app:logit lens} for formal details). We quantify this emergence on the object-centric visual tokens $\mathcal{I}_O$ (identified in Circuit~\ding{192}) using two metrics relative to the target concept $s_c$:
\begin{enumerate}[leftmargin=*, noitemsep, topsep=0pt]
    \item \textit{Correspondence Rate ($C_R^{(\ell)}$).} A hard metric measuring the fraction of tokens whose top-ranked vocabulary projection matches the target concept:
    \begin{equation}
    \small
    C_R^{(\ell)} \triangleq \frac{1}{|\mathcal{I}_{O}|} \sum_{i\in\mathcal{I}_{O}}
    \mathbb{I}\!\left(\arg\max_{w \in \mathcal{V}} P(w \mid h_i^{(\ell)}) \in \mathcal{T}(s_c)\right).
    \end{equation}
    \vspace{-0.1in}
    \item \textit{Answer Probability ($A_P^{(\ell)}$).} A soft metric capturing the average confidence (probability mass) assigned to the target concept:
    \begin{equation}
    \small
    A_P^{(\ell)} \triangleq \frac{1}{|\mathcal{I}_{O}|} \sum_{i\in\mathcal{I}_{O}} P(s_c \mid h_i^{(\ell)}).
    \end{equation}
    \vspace{-0.1in}
\end{enumerate}




\begin{figure}[t]
    \centering
    \includegraphics[width=1\linewidth]{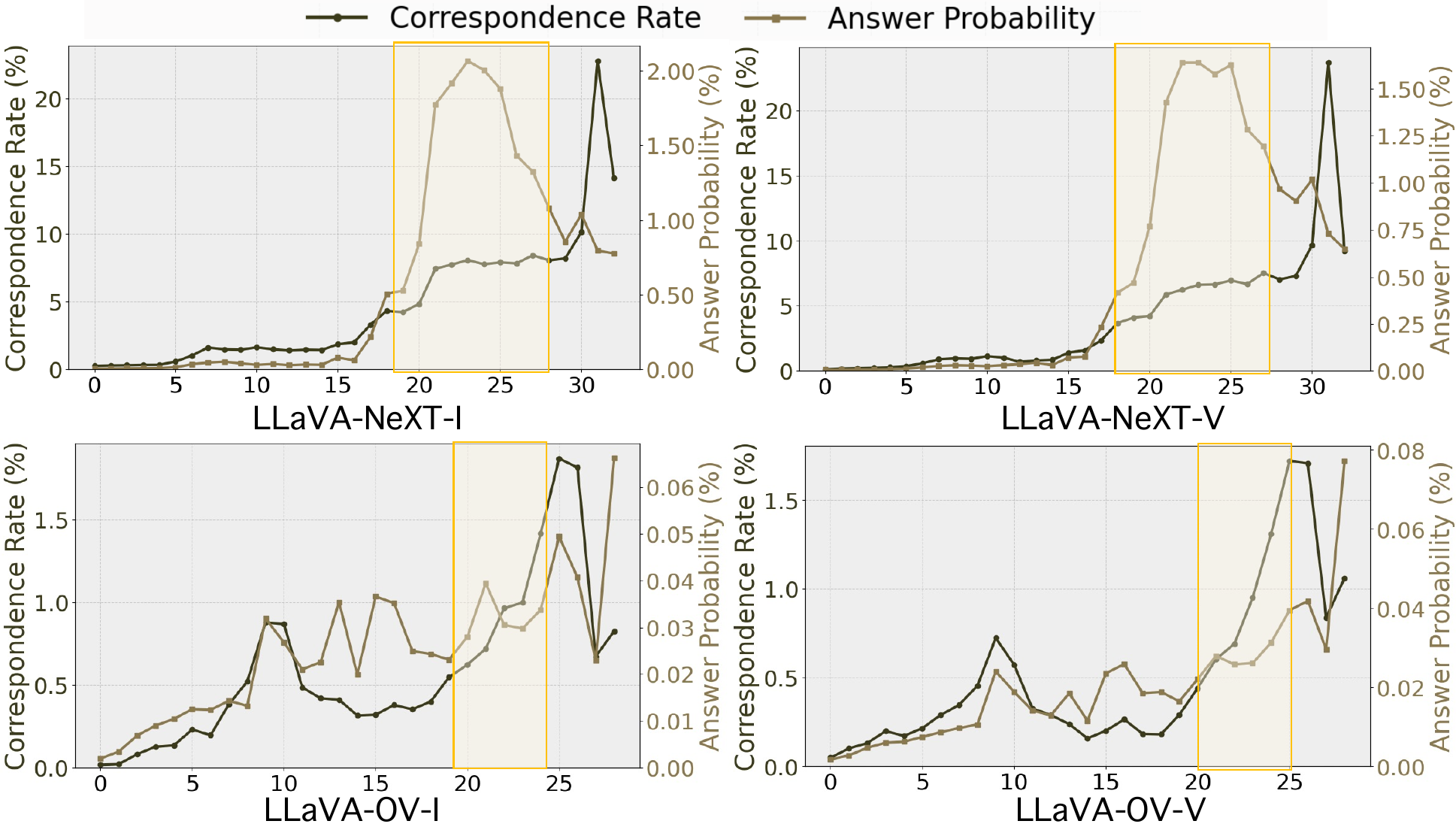}
    \vspace{-0.15in}
    \caption{
        Quantitative analysis of semantic tracing. 
        Both metrics show a sharp increase in the mid-to-late layers, indicating that abstract semantic concepts are consolidated deep within the network.
    }
    \vspace{-0.15in}
    \label{fig:semantic_tracing}
\end{figure}
%
%
Together, $C_R^{(\ell)}$ and $A_P^{(\ell)}$ characterize both the reliability and the strength with which object or action semantics are expressed in the language space at each layer.
\subsubsection{When do video-conditioned semantics become language-aligned?}
\figref{fig:semantic_tracing} plots $C_R^{(\ell)}$ and $A_P^{(\ell)}$ as a function of layer depth.
Across models and concepts, both metrics remain low in early layers, despite the presence of object-aligned visual tokens in the input. 
As depth increases, we observe a sharp and consistent rise in both correspondence rate and answer probability within the mid-to-late layer interval, after which the curves plateau.

This behavior indicates that projection alone does not render visual information immediately usable for language decoding. 
Instead, object and action semantics undergo a depth-dependent transformation inside the LLM, becoming stably expressed in the vocabulary space only after sufficient processing.
Importantly, this transition is substantially more pronounced for object-aligned tokens $\mathcal{I}_O$ than for control token sets, indicating that the observed effect reflects video-conditioned semantic consolidation rather than a generic property of the residual stream. The attention knockout and scaling experiments presented in \appref{app:Circuit2} further support our observations.
\begin{figure*}[t]
    \centering
    \includegraphics[width=.9\linewidth]{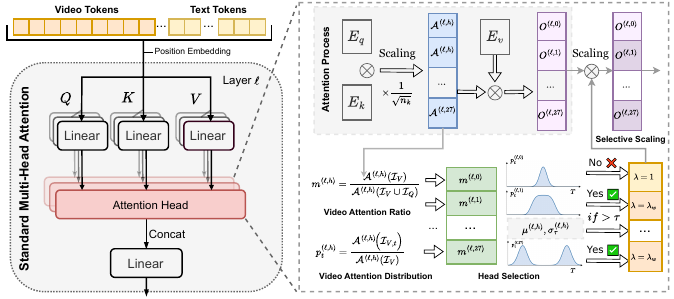}
    \caption{\textbf{Surgery  framework.} We first identify \textit{temporal heads} that exhibit both high video attention and broad temporal dispersion. We then intervene by amplifying these specific heads during the decoding phase. This process validates the causal role of these computational units in handling long-range temporal dependencies.}
    \label{fig:surgery}
\end{figure*}
\begin{table*}[ht]
\centering
\caption{\textbf{Temporal Robustness Analysis.} 
Results are reported as the \textbf{Accuracy (\%)} on the correct-only subset across three temporal perturbation types: \textsc{Reverse}, \textsc{Local Swap}, and \textsc{Shuffle}. 
Higher values indicate better stability against temporal perturbations. 
\textcolor{red}{$\downarrow \Delta$} denotes the performance drop relative to the unperturbed baseline.}
\vspace{-0.05in}
\label{tab:temporal_perturbation_transposed}
\resizebox{\textwidth}{!}{%
\begin{tabular}{l | ccc | ccc | ccc}
\toprule
\multirow{2}{*}{\textbf{Model}} & \multicolumn{3}{c}{\textsc{Reverse}} & \multicolumn{3}{c}{\textsc{Local Swap}} & \multicolumn{3}{c}{\textsc{Shuffle}} \\
\cmidrule(lr){2-4} \cmidrule(lr){5-7} \cmidrule(lr){8-10}
 & Interaction & Sequence & Avg & Interaction & Sequence & Avg & Interaction & Sequence & Avg \\
\midrule
LLaVA-OneVision-OV & \val{52.9}{47.1} & \val{57.0}{43.0} & \val{54.4}{45.6} & \val{96.4}{3.6} & \val{98.2}{1.8} & \val{97.0}{3.0} & \val{75.5}{24.5} & \val{85.5}{14.5} & \val{79.3}{20.7} \\
LLaVA-OneVision-SI & \val{54.1}{45.9} & \val{52.6}{47.4} & \val{53.6}{46.4} & \val{96.6}{3.4} & \val{98.7}{1.3} & \val{97.4}{2.6} & \val{79.7}{20.3} & \val{87.0}{13.0} & \val{82.4}{17.6} \\
Qwen 2.5 VL        & \val{80.7}{19.3} & \val{77.3}{22.7} & \val{79.4}{20.6} & \val{88.3}{11.7} & \val{92.0}{8.0} & \val{89.6}{10.4} & \val{87.2}{12.8} & \val{92.4}{7.6}  & \val{89.1}{10.9} \\
\bottomrule
\end{tabular}%
}
\vspace{-0.1in}
\end{table*}
\subsubsection{How does temporal structure affect semantic consolidation?}
To understand how temporal evidence contributes to this consolidation process, we repeat the same semantic tracing procedure under controlled temporal perturbations of the input video, including frame shuffling and reversal. 
These interventions preserve per-frame visual content while disrupting temporal structure.

As shown in \tabref{tab:temporal_perturbation_transposed}, temporal perturbations degrade downstream performance, with stronger global disruptions (e.g., reversal) causing substantially larger drops than mild local reordering, indicating that temporal structure is behaviorally relevant.
%
%
%
This suggests that temporal information influences decoding primarily through mechanisms operating during semantic consolidation, rather than by altering early visual representations.
\subsubsection{What does semantic tracing reveal about temporal reasoning?}
\begin{figure}[htbp]
    \centering
    \includegraphics[width=0.9\linewidth]{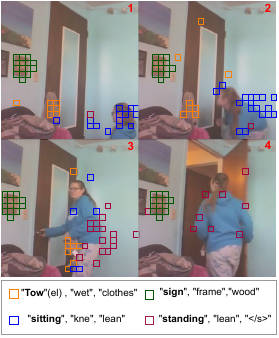}
    \caption{Qualitative example of semantic tracing illustrating how the model captures \textit{temporal dynamics}. 
    The sequence from (a) to (d) shows the evolution of the top-3 most frequent word groups for the single token position. 
    %
    %
    }
    \label{fig:qual}
    \vspace{-0.15in}
\end{figure}

Beyond aggregate layer-wise metrics, we inspect how decoded concepts evolve over time at fixed token positions.
As shown in \figref{fig:qual}, the decoded concepts at a fixed token position evolve in tandem with the instantaneous visual appearance of each frame.
In the standing-up example, ``sitting''-related tokens persist into frames where the action has already begun, and only gradually give way to ``standing''-related tokens.
This lagged, per-frame semantic update is consistent with a static, frame-by-frame description strategy: the model tracks momentary visual states but does not robustly encode the transition itself as a coherent temporal relation.
Taken together with the strong degradation under temporal reversal, these behaviors suggest that temporal understanding in current LVLMs relies on implicit evidence accumulation and recency-biased aggregation, rather than explicit modeling of state transitions. More interactive examples are provided in the \appref{app:circuit2 example}, along with an anonymous website demo for exploration.
\begin{tcolorbox}[
  colback=black!5,
  colframe=black!70,
  title=\textbf{Circuit~\ding{193} Takeaway},
  fonttitle=\bfseries,
  boxrule=0.6pt,
  arc=2mm,
  left=6pt,
  right=6pt,
  top=6pt,
  bottom=6pt,
  before skip=6pt,
  after skip=6pt
]
Video-conditioned object and action concepts become language-aligned only within the mid-to-late semantic consolidation interval, where evidence is integrated across frames.
\end{tcolorbox}

\section{\algname-derived Surgery}\label{sec:surgery_recovery}
Circuits~\ding{192} and~\ding{193} identify two prerequisites on effective temporal reasoning in LVLMs: 
(1) task-critical semantics enter the model through object-aligned video tokens, and 
(2) these semantics become language-aligned only within a mid-to-late consolidation internal.

Building on these observations, we design a surgical recovery method turning these observations into a targeted intervention.
%
%
Specifically, we proceed in two stages:
first identifying such \textit{temporal attention heads} using
interpretable temporal routing statistics, 
and then selectively amplifying their contributions to test whether they causally support temporal understanding.
\subsection{Identifying Temporal Attention Heads}
\label{sec:identifying_temporal_heads}
The findings in Circuit~\ding{193} suggest that temporal reasoning does not emerge from early visual processing, but instead depends on how evidence from multiple frames is aggregated once concepts become language-aligned. 
This motivates a concrete hypothesis: only a small subset of attention heads actively integrate information across time in a way that can influence decoding.
To test this hypothesis, we model each attention head as a two-stage routing mechanism. 
A head must first decide whether to attend to video tokens at all, and then decide how to distribute that attention across temporal slices.
We utilize TempCompass~\citep{liu2024tempcompass} as the primary benchmark to validate the effectiveness of our intervention on temporal reasoning.

\paragraph{Video routing.}
We first quantify whether an attention head meaningfully consumes video evidence.
Heads with negligible video routing are unlikely to participate in temporal reasoning and are excluded from further consideration.
Following the notation in \secref{sec:method}, let $q$ denote a query token position, and let $\alpha^{(\ell,h)}_{q \to j}$ be the attention weight from $q$ to token $j$ at layer $l$, head $h$, we define the attention mass routed from $q$ into a token subset $\mathcal{S}$ as $\mathcal{A}^{(\ell,h)}_q(S)\;\triangleq\;\sum_{j\in S}\alpha^{(\ell,h)}_{q\to j}.$
We partition token indices into video tokens $\mathcal{I}_V=\bigcup_{t=1}^T \mathcal{I}_{V,t}$, grouped by frame (or temporal slice) $t$, and non-video tokens $\mathcal{I}_Q$.
Using this partition, we measure how strongly head $(\ell, h)$ routes attention to the video stream via the \textit{video attention ratio}, which we define as:
\begin{equation}
m^{(\ell,h)} \triangleq
\frac{\mathcal{A}^{(\ell,h)}_q(\Iv)}{\mathcal{A}^{(\ell,h)}_q(\Iv \cup \Iq) + \varepsilon}.
\label{eq:video_attention_ratio}
\vspace{-0.1in}
\end{equation}
Intuitively, $m^{(\ell,h)}$ serves as a gating score.
It is close to zero when the head largely ignores video tokens, and increases as the head allocates a greater ratio of its attention capacity to video-derived evidence.
In the following, we retain only heads with sufficiently large video attention ratio, ensuring that subsequent temporal analysis focuses on mechanisms that actually ingest visual information.

\paragraph{Temporal distribution across frames.}
Conditioned on attending to video tokens, we next characterize how each head distributes its attention across frames.
Intuitively, a head that only attends to a single frame--even if it attends strongly--cannot integrate temporal evidence. 
In contrast, a head that draws information from multiple frames can better support temporal aggregation.

To make this notion precise, we define a frame-wise (slice-wise) attention distribution.
Let $\mathcal{I}_{V,t}$ denote the set of video tokens corresponding to frame (or temporal slice) $t$. 
We define the normalized attention mass assigned by head $(\ell,h)$ to slice $t$ as $ p^{(\ell,h)}_t \triangleq
\frac{\mathcal{A}^{(\ell,h)}_q\!\big(\mathcal{I}_{V, t}\big)}{\mathcal{A}^{(\ell,h)}_q(\mathcal{I}_{V}) + \varepsilon},
\qquad \sum_{t=1}^{T} p^{(\ell,h)}_t = 1.
\label{eq:slice_distribution}
$
This slice distribution makes temporal specialization interpretable: a head that locks onto a single frame yields a highly peaked $p_t$, whereas genuine cross-frame integration yields a broader distribution.

To quantitatively characterize temporal specialization, we measure both the \textit{breadth} of the slice distribution (how many frames receive non-trivial attention) and its \textit{temporal span} (how far that attention extends along time).
We first quantify how many temporal slices contribute non-trivially using the effective number of slices defined via the Hill number of order~2 (inverse Simpson index)~\citep{somerfield2008simpson}:
\begin{equation}
n_{\mathrm{eff}}^{(\ell,h)} \triangleq
\frac{1}{\sum_{t=1}^{T} \left(p^{(\ell,h)}_t\right)^2 + \varepsilon}.
\label{eq:effective_slices}
\end{equation}
which captures the breadth of temporal support: $n_{\mathrm{eff}} \approx 1$ indicates near-exclusive focus on a single slice, while larger values indicate that attention is distributed across frames.

To capture \emph{how far} evidence spans along time, we compute the second central moment of the slice distribution.
Using normalized slice indices $\tau_t \triangleq \frac{t-1}{T-1}\in[0,1]$ (for comparability across different $T$), we define:
\begin{equation}
\small
\sigma_{\tau}^{(\ell,h)} =
\sqrt{\sum_{t=1}^{T} p^{(\ell,h)}_t \, \left(\tau_t - \mu^{(\ell,h)}\right)^2 },
\text{ where }
\mu^{(\ell,h)} = \sum_{t=1}^{T} p^{(\ell,h)}_t \, \tau_t.
\label{eq:temporal_std}
\end{equation}
While $n_{\mathrm{eff}}$ measures the breadth of the support, standard deviation $\sigma_{\tau}$ distinguishes local multi-slice smoothing from
long-range temporal integration.

\paragraph{Temporal head selection.}
We designate a head $(\ell,h)$ as \textit{temporal} only if it consistently (i) routes non-trivial attention mass to the video stream, (ii) draws support from multiple frames, and (iii) spans a sufficiently wide temporal range.
Concretely, we require
\begin{equation}
\bar{m}^{(\ell,h)} \ge \tau_m,\quad
\overline{n_{\mathrm{eff}}}^{(\ell,h)} \ge \tau_{\mathrm{eff}},\quad
\bar{\sigma}_{\tau}^{(\ell,h)} \ge \tau_{\sigma},
\label{eq:temporal_head_thresholds}
\end{equation}
%
All thresholds $(\tau_m,\tau_{\mathrm{eff}},\tau_{\sigma})$ and metric ablations are further specified in \appref{app:surgery example}, and the full selection pipeline is illustrated in \figref{fig:surgery}.

\subsection{Temporal Head Amplification}
Having identified a sparse set of temporally specialized heads $\mathcal{S}_{\text{temp}}$ using the routing and dispersion criteria in \eqnref{eq:temporal_head_thresholds}, we now test whether strengthening these heads causally improves temporal understanding.
Guided by Circuit~\ding{193}, which localizes the semantic consolidation interval, we restrict amplification to the critical layer range $\mathcal{L}_{\text{crit}}$ (typically mid-to-late layers) where video-conditioned concepts become language-aligned and most likely to influence decoding.
For an attention head $h$ at layer $\ell$, let $\mathbf{O}^{(\ell,h)} \in \mathbb{R}^{d_k}$ denote its output vector after attention but before the final linear projection $\mathbf{W}_O$.
We apply a scaling factor $\lambda$ to amplify the contribution of identified temporal heads, i.e.
\begin{equation}
\tilde{\mathbf{O}}^{(\ell,h)} = 
\begin{cases} 
\lambda \cdot \mathbf{O}^{(\ell,h)}, & \text{if } (\ell,h) \in \mathcal{S}_{\text{temp}} \text{ and } \ell \in \mathcal{L}_{\text{crit}} \\
\mathbf{O}^{(\ell,h)}, & \text{otherwise}
\end{cases}\nonumber
\label{eq:intervention}
\end{equation}
where $\alpha$ is a hyperparameter controlling the intervention strength.
Crucially, this operation amplifies the \textit{magnitude} of the temporal evidence routed into the residual stream without altering the distribution of the attention weights itself, thereby preserving the head's intrinsic routing logic while enhancing its impact on the final generation.
\begin{figure}[htbp]
    \centering
    \includegraphics[width=.8\linewidth]{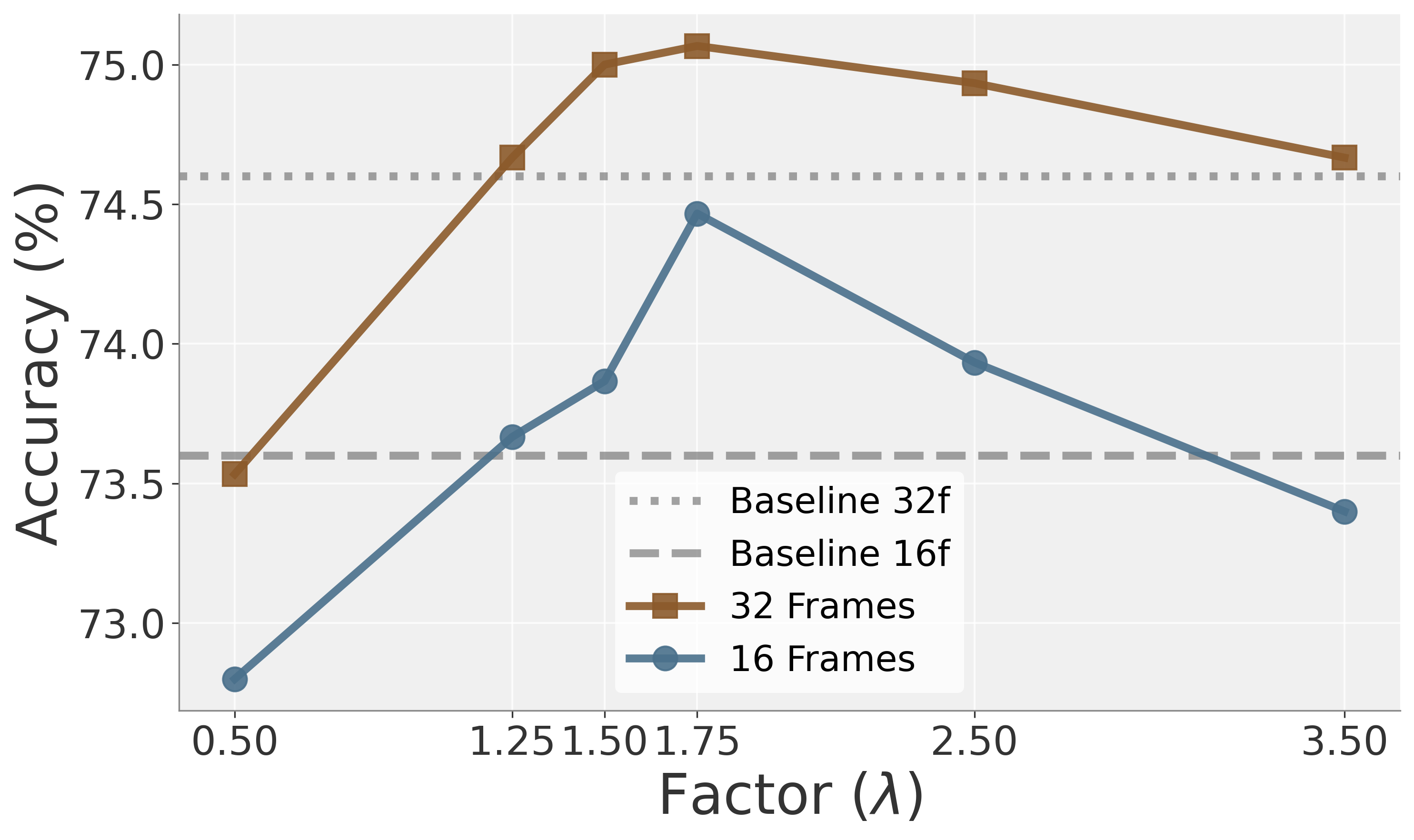}
    \caption{Sensitivity analysis of the head amplification factor $\lambda$ across layer windows.}
    \label{fig:sensitivity}
\end{figure}
\vspace{-0.2in}
\subsection{Experiments and Validation of Findings}
%
\paragraph{Experiment results.}
As shown in \tabref{tab:temporal_head_ablation}, amplifying the identified temporal heads improves performance on TempCompass~\citep{liu2024tempcompass} across all evaluated LVLMs, with gains reaching up to +2.4\% absolute accuracy. 
These improvements are observed under both 16-frame and 32-frame settings, indicating that the intervention benefits genuine temporal reasoning rather than exploiting a specific input length or configuration.
Notably, these gains are achieved without modifying model parameters or retraining, demonstrating that temporal reasoning capacity already exists within the model and can be selectively strengthened at inference time, with the optimal scaling factor $\alpha$ for these heads further illustrated in Fig.~\ref{fig:sensitivity}.

\textbf{Validation of \algname findings.}
\tabref{tab:temporal_head_ablation} shows that gains concentrate in \emph{mid-to-late} layers: injecting into earlier layers yields negligible improvements or even slight degradation, while later blocks (e.g., Layers 20--25/25--30) produce the most reliable uplift across models and frame counts.
This layer locality is consistent with observation in \figref{fig:semantic_tracing}: semantic concepts become strongly decodable only in mid-to-late layers, implying that temporal evidence routing is most effective once object semantics have been consolidated into language-aligned representations.
In other words, early layers primarily transmit and normalize visual features, whereas later layers are where cross-slice evidence aggregation can directly influence answer tokens. More qualitative examples can be found in \appref{app:surgery qualitive examples}, which further demonstrate the robustness of our findings.

\begin{table}[t]
\centering
\caption{\textbf{Effectiveness of Temporal Head Amplification.} We evaluate the proposed inference-time intervention on the TempCompass benchmark under 16- and 32-frame settings. Results are reported without retraining, following the protocols in \secref{subsec:prelim_setup}.}
\label{tab:temporal_head_ablation}
\resizebox{\linewidth}{!}{
\begin{tabular}{l cc cc cc}
\toprule
& \multicolumn{2}{c}{\textbf{Qwen 2.5 VL}} & \multicolumn{2}{c}{\textbf{LLaVA-OV-V}} & \multicolumn{2}{c}{\textbf{LLaVA-OV-I}} \\
\cmidrule(lr){2-3} \cmidrule(lr){4-5} \cmidrule(lr){6-7}
\textbf{Injection Layers} & $T=16$& $T=32$ & $T=16$ & $T=32$ & $T=16$ & $T=32$ \\
\midrule
\textit{Baseline (None)} & 73.6 & 74.6 & 72.2 & 72.2 & 63.6 & 63.6 \\
\midrule
Layers 0--5   & \loss{73.6}{0.0}     & \loss{74.2}{0.4} & \loss{72.0}{0.2} & \loss{71.4}{0.8} & \gain{63.8}{0.2} & \loss{63.6}{0.0} \\
Layers 5--10  & \loss{73.4}{0.2} & \loss{74.4}{0.2} & \loss{72.2}{0.0} & \gain{72.4}{0.2} & \loss{63.6}{0.0}   & \gain{63.8}{0.2} \\
Layers 10--15 & \loss{73.2}{0.4} & \loss{74.2}{0.4} & \gain{72.4}{0.4} & \loss{72.2}{0.0}   & \loss{63.2}{0.4} & \gain{63.8}{0.2} \\
Layers 15--20 & \loss{73.4}{0.2} & \gain{74.8}{0.2} & \loss{71.8}{0.4} & \loss{72.2}{0.0}   & \loss{62.6}{1.0} & \loss{63.2}{0.4} \\
\rowcolor{gray!15} 
Layers 20--25 & \gain{76.0}{\textbf{2.4}} & \gain{75.4}{0.8} & \gain{72.2}{0.0} & \gain{72.6}{0.4} & \loss{62.8}{0.8} & \loss{62.6}{1.0} \\
\rowcolor{gray!15} 
Layers 25--30 & \gain{74.0}{0.4} & \gain{75.0}{0.4} & \gain{73.0}{\textbf{1.0}} & \loss{72.2}{0.0}   & \loss{63.0}{0.6} & \loss{63.2}{0.4} \\
\bottomrule
\end{tabular}
}
\vspace{-0.25in}
\end{table}

\section{Conclusion and Limitation}
We propose \algname to mechanistically dissect video LVLMs by (i) \emph{Visual Auditing} that localizes task-critical object-aligned video tokens and causally verifies their necessity (text-label injection only partially recovers), and (ii) \emph{Semantic Tracing} that shows video semantics become reliably language-aligned mainly in mid-to-late layers, where models often behave closer to evidence aggregation (often bag-of-frames-like) than explicit temporal-state modeling. Guided by these findings, we identify temporally dispersed, video-routing attention heads and selectively amplify them within the consolidation interval, yielding consistent inference-time gains on temporal-heavy benchmarks across backbones and frame budgets. However, we do not yet systematically quantify potential side effects such as calibration shifts or robustness under distribution shift. Finally, our evaluation covers a limited set of model families, and extending the analysis and interventions to broader architectures and larger models remains important future work.

\pagebreak
\newpage
\section*{Impact Statement}
This work aims to enhance our understanding of temporal reasoning in LVLMs by dissecting their ability to interpret and process temporal information across frames. Through detailed analysis of perturbation behaviors, we uncover how LVLMs rely on temporal cues and the limitations they face in modeling complex temporal dynamics. By proposing targeted interventions like temporal head identification and amplification, we seek to advance the design of LVLMs that can robustly handle temporal dependencies, ensuring more accurate and context-aware multimodal understanding. We advocate for the integration of temporal reasoning as a key aspect of model alignment, urging the research community to prioritize this alongside other capabilities, fostering the development of more reliable, robust, and ethically sound AI systems.
\bibliography{custom,Multi-modal_LLM,VideoLLM,main,Trustworthy}
\bibliographystyle{icml2026}

\newpage
\appendix
\onecolumn
\clearpage
\appendix
\onecolumn
\section*{Appendix}
\section{Analysis Setup}\label{app:Analysis Setup}
\subsection{Question Probing Setup}
We use two complementary question formats that probe object and action understanding to evaluate the impact of token-level interventions under controlled decoding conditions, 
\begin{enumerate}[leftmargin=*, noitemsep, topsep=0pt]
    \item \textit{Open-ended questions.} 
    We formulate specific questions targeting objects and actions depicted in the video (e.g., ``Which object did the person throw in the video?''). 
    These questions are designed to be answerable solely based on the provided video key frames. 
    To ensure controlled comparison across ablation conditions, we pre-fill the model's response with a fixed prefix (e.g., ``The object is") and analyze the next generated token before and after intervention.
    \item \textit{Close-ended questions.} 
    We prompt the model with binary questions (e.g., ``Is there a/an [object] in this video?"), and assess the impact of intervention by comparing the model's next-token prediction. 
    A flip from ``Yes'' to ``No'' after the intervention serves as evidence that the manipulation was crucial for the model's reasoning and understanding of the input video.
\end{enumerate}
\subsection{Frames Visualization}\label{app:Frames}
\begin{figure}[ht]
\centering
\includegraphics[width=.9\linewidth]{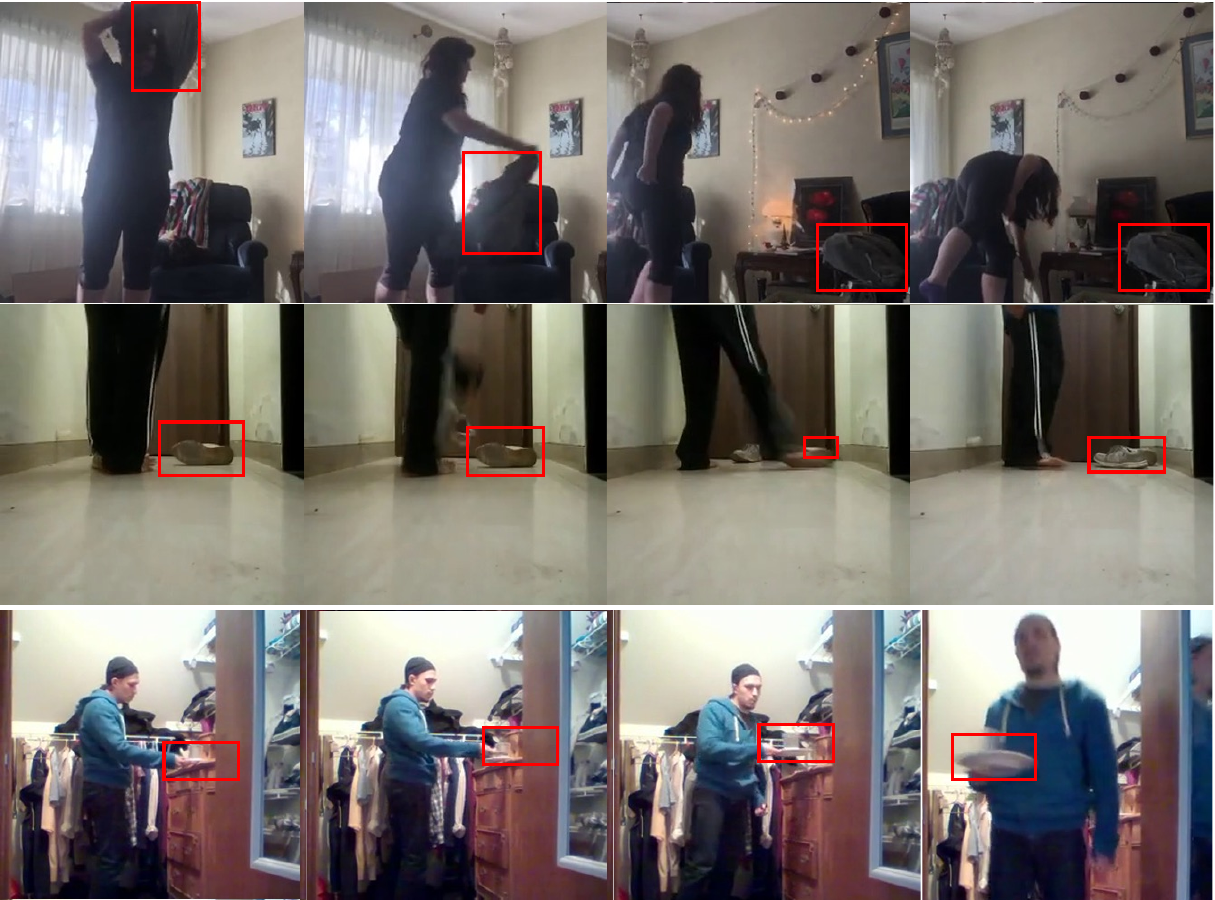}
\caption{
Visualization of Input Keyframes. Each row displays a sequence of frames provided to the model as input for a specific video. The red bounding boxes highlight the ground-truth object pertinent to the task's question (e.g., the object being picked up, kicked, or taken). It is important to note that these bounding boxes are included here for clarity and were not provided to the model during inference.
}
\label{fig:app_frames}
\end{figure}

This section visualizes sample input keyframes from the videos to provide a clearer understanding of the data used in our experiments. Each row in Figure \ref{fig:app_frames} represents a distinct video sequence fed into the LVLMs for analysis. These examples are representative of the scenarios in our dataset, encompassing a variety of everyday actions, objects, and environments. By visualizing the raw inputs, we aim to illustrate the visual complexities, such as changes in viewpoint, object scale, and partial occlusions, that the model must handle to perform accurate semantic reasoning.

\newpage
\section{Supplementary Experiments and Details: Circuit~\ding{192} and~\ding{193}}\label{app:Circuit2}
\subsection{Logit lens Setup}\label{app:logit lens}
Our analysis focuses exclusively on the internal representations of projected visual tokens as they propagate through the LLM backbone, before any autoregressive generations. 
Let $\mathcal{I}_O$ denote the set of object-centric visual token indices identified in Circuit~\ding{192}, and let $n_O \triangleq |\mathcal{I}_O|$. 
For each token position $i \in \mathcal{I}_O$, we denote by $h_i^{(\ell)} \in \mathbb{R}^d$ the hidden state at layer $\ell$. 
At each layer, we project this hidden state into the vocabulary space using the frozen language-model output projection $W_{LM}$, defining a layer-wise token distribution
\begin{equation}
    P(\cdot \mid h_i^{(\ell)}) = \mathrm{softmax}(W_{LM} h_i^{(\ell)}).
\end{equation}
For each sample, we define a target semantic concept $s_c$ corresponding to the ground-truth object or action referenced by the question (e.g., ``\texttt{microwave}''). 
If $s_c$ is tokenized into multiple sub-word tokens, we define the target token set $\mathcal{T}(s_c)\subseteq\mathcal{V}$ accordingly, and use 
\begin{equation}
P(s_c \mid h_i^{(l)})\triangleq \sum_{w\in\mathcal{T}(s_c)} P(w\mid h_i^{(\ell)}).
\end{equation}
When $s_c$ maps to a single token, $\mathcal{T}(s_c)=\{w_c\}$ and the definition reduces to $P(w_c \mid h_i^{(\ell)})$.

\subsection{Scaling Experiments}\label{app:Scaling}
 We conducted supplementary experiments to verify that our conclusions generalize to larger-scale models. We replicated our core analyses on the LLaVA-NeXT-34B model variants, with results that closely mirror those presented in the main body of the paper. The visual token ablation study on the 34B models reaffirms the principle of spatial localization for semantic information. Ablating object-specific tokens incurs significantly more substantial performance degradation than removing larger quantity of random tokens (in \tabref{tab:attention_knockout_enhanced_34b}). 


Furthermore, our semantic tracing analysis on the 34B architecture, depicted in \figref{fig:prob_heatmap_34b}, reveals a conceptual emergence pattern consistent with our earlier observations. Both the Correspondence Rate and Answer Probability remain negligible through the initial layers before exhibiting a sharp, concurrent rise beginning around layer 40. This trend indicates that abstract, language-aligned concepts are consolidated in the deeper layers of the network, irrespective of model scale. These scaling experiments provide robust evidence that the mechanisms of semantic localization and late-stage conceptual formation are fundamental properties of the tested LVLM architectures.
\begin{minipage}[!t]{\linewidth}
\begin{minipage}[!t]{0.53\linewidth}
\centering
\renewcommand{\arraystretch}{1.1}
\scriptsize
\tabcolsep=1.00mm
\begin{tabular}{@{}c c cc cc@{}}
\toprule
\textbf{Ablation}& \textbf{Tokens} & \multicolumn{2}{c}{\textbf{LLaVA-NeXT-34B-I}} & \multicolumn{2}{c}{\textbf{LLaVA-NeXT-34B-V}} \\
\cmidrule(lr){3-4} \cmidrule(lr){5-6}
\textbf{Type} & \textbf{Number} & \multicolumn{1}{c}{Open} & \multicolumn{1}{c}{Close} & \multicolumn{1}{c}{Open} & \multicolumn{1}{c}{Close} \\
\midrule
\multicolumn{6}{@{}c}{\textit{Control Groups (Low Tokens)}} \\
Baseline& 0 & \perf{0} & \perf{0} & \perf{0} & \perf{0} \\
Register & 13 & \perf{44.68} & \perf{3.14} & \perf{44.68} & \perf{98.2} \\
\midrule
\multicolumn{6}{@{}c}{\textit{Object-based Ablation}} \\
 & 304 & \perf{90.75} & \perf{70.16} & \perf{85.37} & \perf{88.35} \\
Object & 413 & \perf{94.24} & \perf{78.88} & \perf{89.32} & \perf{88.72} \\
 & 573 &  \perfmax{94.24} & \perfmax{82.02} & \perfmax{89.74} & \perfmax{88.21} \\
\midrule
\multicolumn{6}{@{}c}{\textit{Control Groups (High Tokens)}} \\
& 100 & \perf{45.72} & \perf{3.14} & \perf{6.92} & \perf{70.26} \\
Random & 350 & \perf{44.5} & \perf{3.49} & \perf{11.54} & \perf{68.97} \\
& 500 & \perf{45.55} & \perf{3.66} & \perf{16.92} & \perf{69.23}  \\
& 900 &  \perf{40.14} & \perf{3.17} & \perf{21.67} & \perf{70.33} \\
\bottomrule
\end{tabular}
\makeatletter\def\@captype{table}\makeatother\caption{Accuracy (\%) from visual token ablation on question-answering performance across 34B models. The $\downarrow$ symbol indicates the magnitude of this performance drop.}
\label{tab:attention_knockout_enhanced_34b}
\end{minipage}
\begin{minipage}[!t]{0.45\linewidth}
\centering
\includegraphics[width=.85\linewidth]{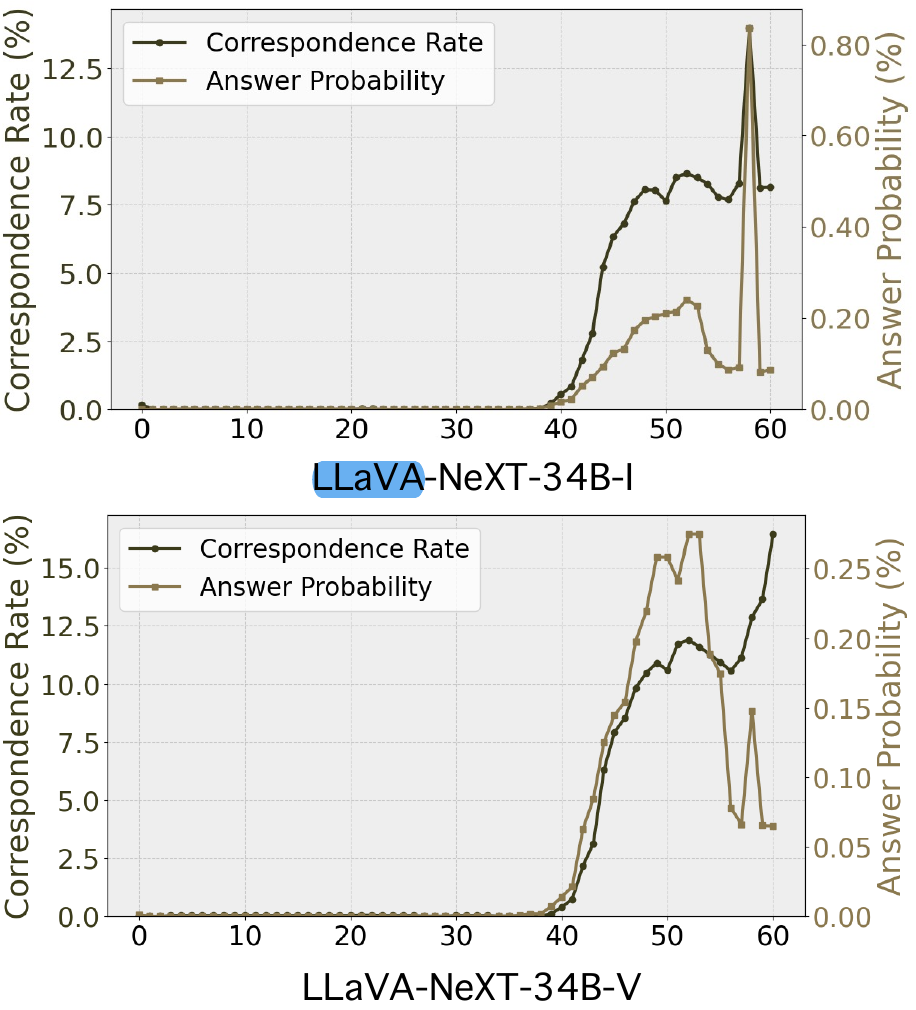}
\makeatletter\def\@captype{figure}\makeatother\caption{Quantitative analysis of semantic tracing on 34B model size. }\label{fig:prob_heatmap_34b}
\end{minipage}
\end{minipage}
\newpage
\subsection{Attention Knockout}
Our earlier token-level interventions establish that task-relevant object semantics are spatially localized in the video embedding (Circuit~\ding{192}). 
However, localization alone does not clarify \emph{how} such evidence is routed through the LLM backbone when producing an answer, nor whether temporal reasoning can be supported by object-centric evidence in isolation.
This appendix section therefore characterizes the \emph{attention flow} that supports the \emph{first generated token}, with the goal of providing auxiliary evidence for the layer-localization claims used in Circuit~\ding{193}.

\paragraph{Setup and masking protocol.}
We conduct all experiments on the \llmname{LLaVA-NeXT} family.
To control information flow during prediction of the first output token, we apply an attention knockout mask that blocks attention edges from a specified source token set to the final answer token, restricted to a chosen layer window.
We partition layers into five windows (Early, Early-to-Mid, Mid, Mid-to-Late, Late), and apply the same mask to \emph{all heads} within the selected window.

We consider two complementary masking regimes:
(i) \textbf{object-centric masking} blocks attention from object-region tokens and buffers (O, O+1, O+2) to the final token; 
(ii) \textbf{contextual masking} blocks attention from all non-object tokens and their complements (C, C+1, C+2) to the final token.
For each configuration, we report (a) \textbf{accuracy drop} (\tabref{tab:attention_knockout_enhanced}) and (b) \textbf{answer probability drop} for the correct token (\figref{fig:prob_heatmap}).
\paragraph{Analysis.}
This attention-flow diagnostic is designed to \emph{triangulate} the conclusions of Circuits~\ding{192}--\ding{193} by asking a different question: 
given that object evidence is spatially localized in the video embedding (Circuit~\ding{192}), \emph{which} information source (object vs.\ context) is actually \emph{consumed} by the LLM backbone, and \emph{where} in depth this consumption is most critical when forming the first answer token.
\tabref{tab:attention_knockout_enhanced} and \figref{fig:prob_heatmap} show a consistent separation between the roles of contextual versus object-centric information for first-token prediction.

\textbf{From localization to utilization.}
Circuit~\ding{192} shows that ablating object-region tokens catastrophically harms object identification, establishing that object semantics are indeed present and localized.
In contrast, \tabref{tab:attention_knockout_enhanced} and \figref{fig:prob_heatmap} show that \emph{blocking access to non-object context} produces a substantially larger degradation than blocking access to the object region when predicting the first token.
Taken together, these results refine the interpretation of Circuit~\ding{192}: although object semantics are necessary and localized, \emph{they are not sufficient for the model’s initial reasoning state}; the model must first read broader contextual evidence to decide \emph{how} to use object information.

\textbf{Depth-specific division of labor.}
The layer-window results provide a depth-resolved counterpart to Circuit~\ding{193}.
Circuit~\ding{193} (logit-lens tracing) indicates that semantically decodable representations emerge sharply in mid-to-late layers.
Consistently, we find that contextual masking is most harmful in early-to-mid windows, while object-centric masking becomes relatively more harmful in mid-to-late windows.
This alignment supports a two-stage computation: (i) early-to-mid layers integrate global/contextual cues to establish a coarse reasoning scaffold, and (ii) mid-to-late layers selectively incorporate fine-grained object evidence as semantic representations become language-aligned, directly shaping the answer distribution.

Overall, attention-flow analysis offers a complementary lens that links
(1) \emph{spatial localization} of object evidence (Circuit~\ding{192}) and 
(2) \emph{depth-wise emergence} of language-aligned semantics (Circuit~\ding{193})
to a concrete, causal \emph{routing} process about which evidence is accessed at which depth during generation.

\begin{minipage}[!t]{\linewidth}
\begin{minipage}[!t]{0.53\linewidth}
\centering
\renewcommand{\arraystretch}{1.1}
\scriptsize
\tabcolsep=1.00mm
\begin{tabular}{llcccccc}
\toprule
\multirow{2}{*}{\textbf{Model}} & {\textbf{Condition }}  & \multirow{2}{*}{\textbf{Early}} & \textbf{Early-} & \multirow{2}{*}{\textbf{Mid}} & \textbf{Mid-} & \multirow{2}{*}{\textbf{Late}} & \textbf{All} \\
 & (From*) &  & \textbf{Mid} &  & \textbf{Late} & & \textbf{Layers} \\
\midrule
\multicolumn{8}{c}{\textit{Object-centric masking}} \\
\midrule
\multirow{3}{*}{LLaVA-NeXT-I} & O & 0.19 & 0.20 & 0.24 & \textbf{0.25} & 0.23 & 0.29 \\
 & O+1 & 0.20 & 0.23 & 0.25 & \textbf{0.28} & 0.27 & 0.35 \\
 & O+2 & 0.23 & 0.23 & 0.26 & \textbf{0.28} & 0.26 & 0.39 \\
\midrule
\multirow{3}{*}{LLaVA-NeXT-V} & O & 0.25 & 0.26 & 0.27 & \textbf{0.30} & 0.27 & 0.32 \\
 & O+1 & 0.25 & 0.26 & 0.29 & \textbf{0.33} & 0.29 & 0.38 \\
 & O+2 & 0.25 & 0.26 & 0.31 & \textbf{0.33} & 0.32 & 0.42 \\
\midrule
\multicolumn{8}{c}{\textit{Contextual masking}} \\
\midrule
\multirow{3}{*}{LLaVA-NeXT-I} & C & 0.82 & \textbf{0.99} & 0.89 & 0.66 & 0.44 & 0.99 \\
 & C+1 & 0.83 & \textbf{0.97} & 0.86 & 0.59 & 0.39 & 0.97 \\
 & C+2 & 0.83 & \textbf{0.97} & 0.86 & 0.54 & 0.39 & 0.97 \\
\midrule
\multirow{3}{*}{LLaVA-NeXT-V} & C & 0.49 & \textbf{0.91} & 0.89 & 0.92 & 0.62 & 0.97 \\
 & C+1 & 0.47 & \textbf{ 0.90} & 0.86 & 0.88 & 0.56 & 0.94 \\
 & C+2 & 0.46 & \textbf{0.90} & 0.83 & 0.87 & 0.52 & 0.92 \\
\bottomrule
\end{tabular}
\makeatletter\def\@captype{table}\makeatother\caption{\textbf{Accuracy drop} for \llmname{LLaVA-NeXT} models across various layer windows after attention knockout. Masking conditions: \textbf{O}, object tokens; \textbf{O+1/2}, object with 1/2  buffer; \textbf{C}, all tokens except the object; \textbf{C+1/2}, all tokens except the object with 1/2 buffer.}
\label{tab:attention_knockout_enhanced}
\end{minipage}
\hspace{0.1in}
\begin{minipage}[!t]{0.4\linewidth}
\centering
\includegraphics[width=.85\linewidth]{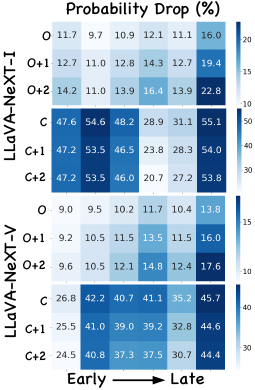}
\makeatletter\def\@captype{figure}\makeatother\caption{\textbf{Answer probability drop} under attention knockout across various layers.}\label{fig:prob_heatmap}
\end{minipage}
\vspace{-0.2in}
\end{minipage}

\subsection{Interactive Examples}\label{app:circuit2 example}
We provide additional qualitative examples to visually illustrate the findings from our circuit-based analysis. These examples showcase the model's process of interpreting video frames to answer specific questions about objects and actions. Each figure includes the input video frames, the posed question, the model's response, and a table showing the semantic evolution of key tokens across different layers, as analyzed through our semantic tracing circuit (Circuit \ding{193}).
More interactive examples in the anonymous interactive \href{https://quaitivesamples.netlify.app/}{demo website}.
\begin{figure}[ht]
 \centering
 \includegraphics[width=0.7\linewidth]{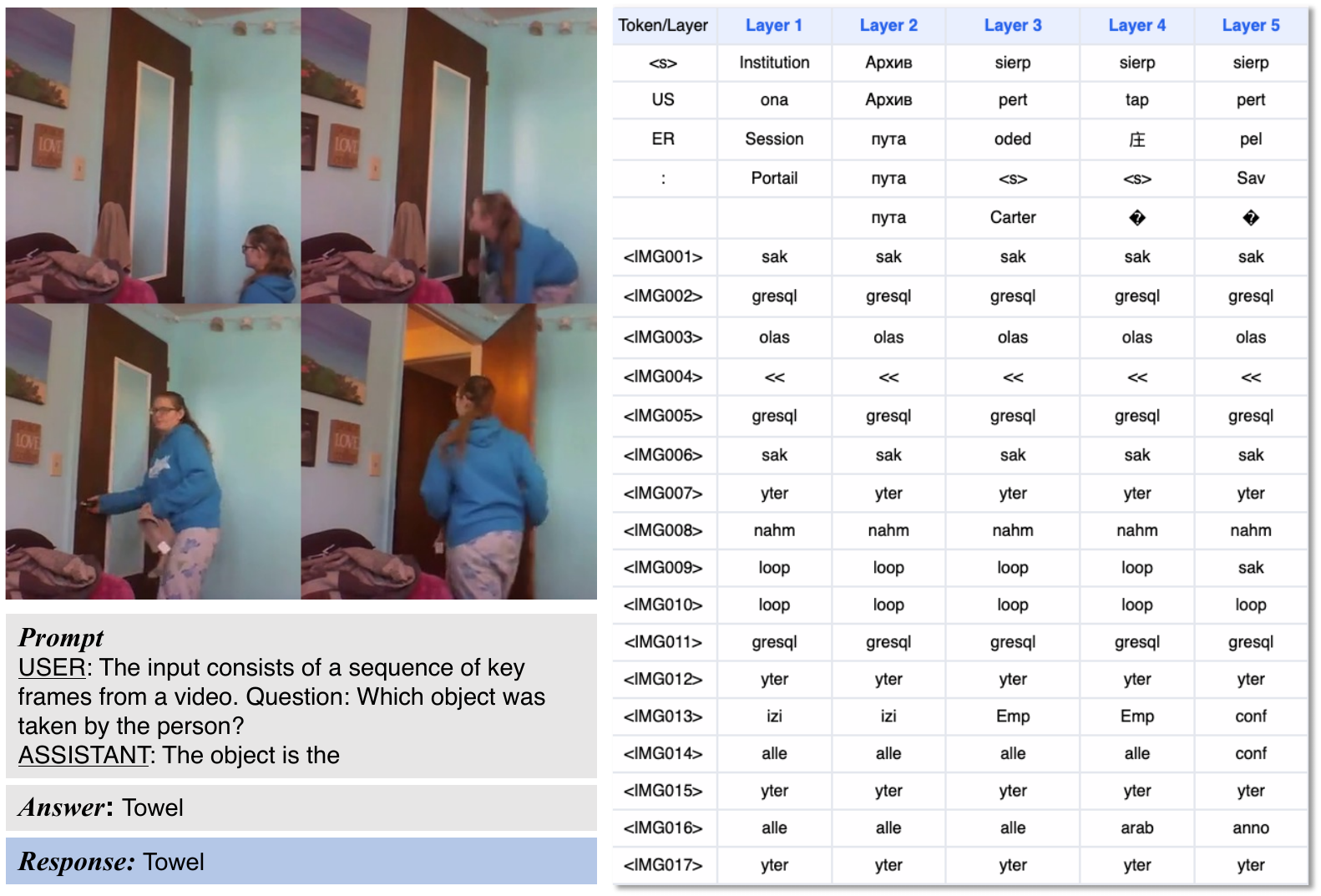}
 \caption{
Qualitative example of the model correctly identifying an object. The user asks which object was taken by the person. The model correctly identifies the "Towel". The accompanying table shows the layer-by-layer semantic tracing for visual and text tokens.
}
\label{fig:app_ex1}
\end{figure}
\clearpage

\begin{figure}[htbp]
\centering
\includegraphics[width=0.7\linewidth]{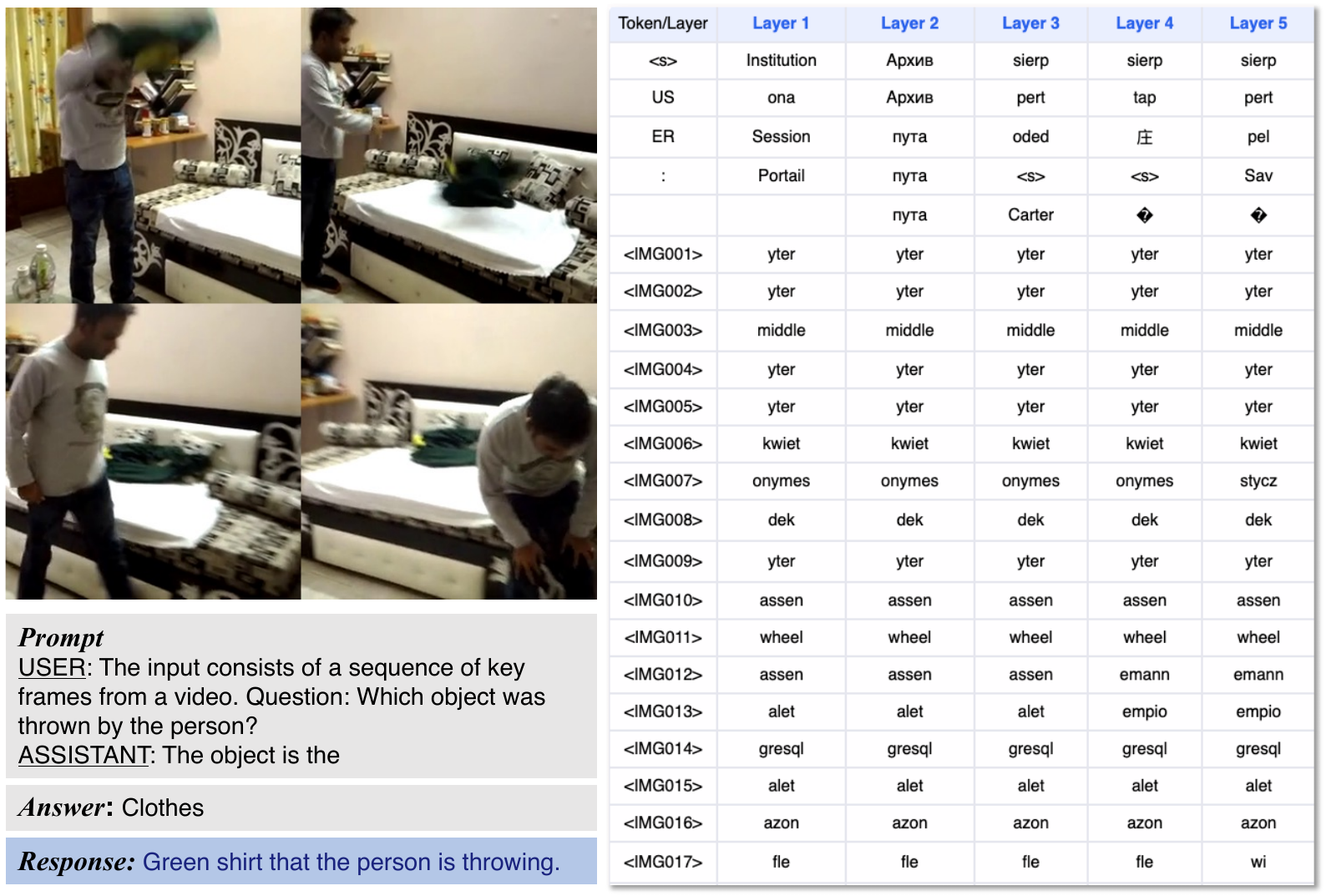}
\caption{
Qualitative example where the model is prompted to identify a thrown object. The model successfully responds that a "Green shirt" was thrown, correctly identifying both the object and its color. The table illustrates the semantic trace, showing how the model processes the visual information through its layers.}
\label{fig:app_ex2}
\vspace{-0.2in}
\end{figure}
\section{Surgery Details and Setup}\label{app:surgery example}
\subsection{Threshold Setting}\label{app:surgery threshold}
This section details the thresholding protocol used to identify \emph{temporal heads} (Eq.~\ref{eq:temporal_head_thresholds}) and the hyperparameters used for head amplification. 
We report all thresholds and aggregation choices to ensure the surgery procedure is fully reproducible.
All thresholds $(\tau_m,\tau_{\mathrm{eff}},\tau_{\sigma},\tau_{\mathrm{occ}})$ are selected on a held-out development split to avoid test leakage.
Concretely, we use a percentile-based calibration to balance (i) head sparsity and (ii) downstream gains under surgery.
Across the explored configurations, we find that the best-performing setting is
$\tau_m=0.1$, $\tau_{\mathrm{eff}}=2.0$, and $\tau_{\sigma}=0.8$,$\tau_{\mathrm{occ}}=0.6$
which achieves the strongest accuracy improvement while maintaining a compact headset.

\subsection{Metric Ablation}\label{app:Metric Ablation}
The head-type ablation in \tabref{tab:temporal_head_layers} shows that partial criteria (e.g., using only $(m, n_{\mathrm{eff}})$ or $(m,\sigma_\tau)$) do not reproduce the full gains, and random head amplification fails to help.
This indicates that the improvement is not explained by a generic increase in attention capacity; rather, it depends on jointly satisfying (i) \emph{video gating} and (ii) \emph{broad, long-range temporal support}.
Together, these results validate our profiling metrics as identifying a functionally meaningful subset of heads that implement time-spanning evidence routing.

\subsection{Qualitive Examples}\label{app:surgery qualitive examples}
\begin{figure}[ht]
\centering
\includegraphics[width=1\linewidth]{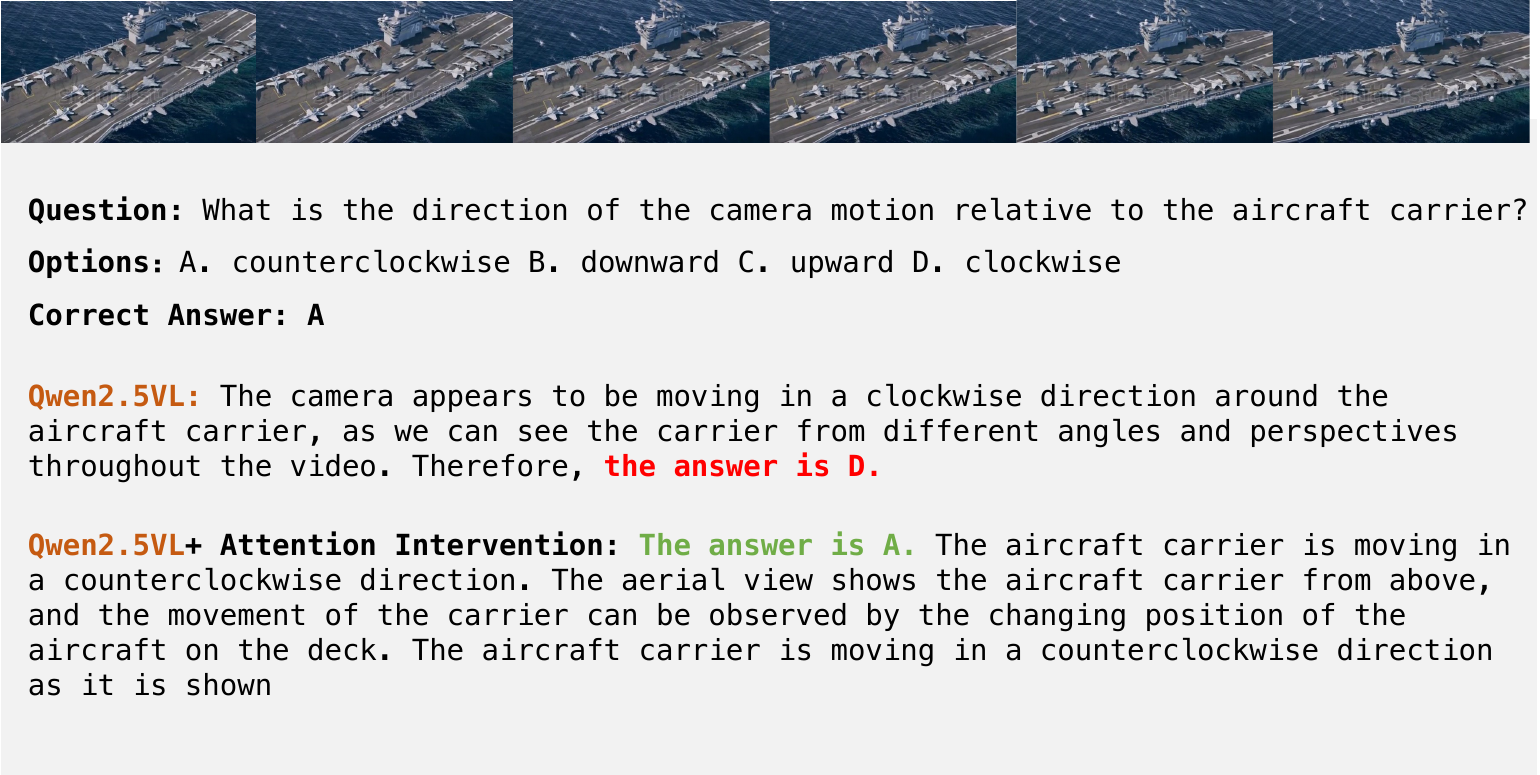}
\caption{\textbf{Qualitative comparison on Visual Dynamics (Camera Motion).} 
The question asks for the direction of camera motion relative to the aircraft carrier. 
The \textbf{Baseline} model  suffers from directional hallucination, incorrectly inferring a \textit{clockwise} motion despite clear visual evidence. 
In contrast, our \textbf{Intervened} model correctly grounds the changing perspective over time, accurately identifying the \textit{counterclockwise} motion (Option A). 
This demonstrates the efficacy of our intervention in rectifying errors related to continuous spatiotemporal changes.}
\label{fig:surgery example1}
\end{figure}

\begin{figure}[ht]
\centering
\includegraphics[width=1\linewidth]{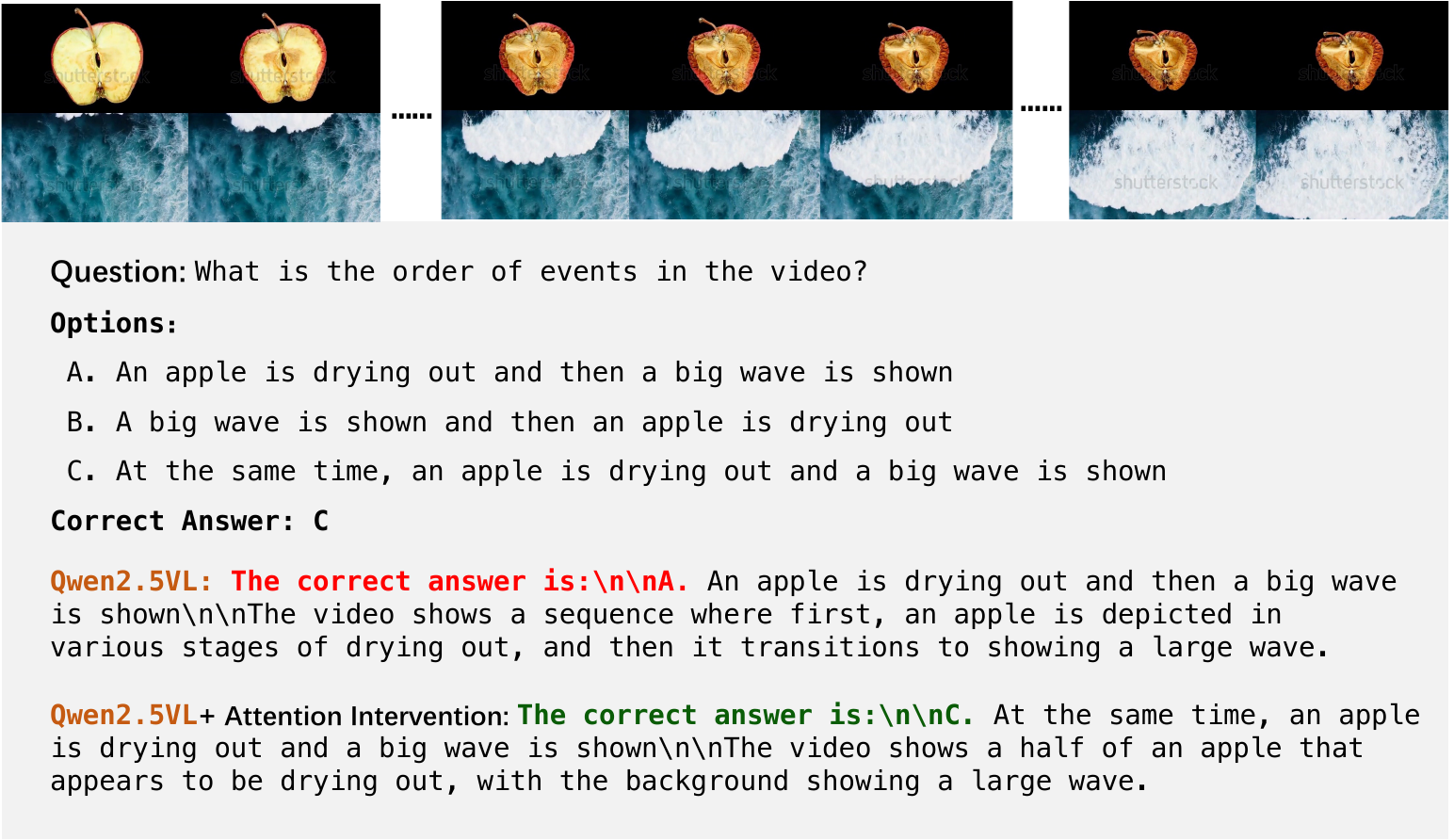}
\caption{\textbf{Qualitative comparison on the Sequence Task (Event Ordering).} 
The question probes the chronological order of two visual events: an apple drying out and a wave. 
The \textbf{Baseline} fails to recognize the split-screen simultaneity, hallucinating a sequential transition (\textit{"...and then a big wave is shown"}). 
The \textbf{Intervened} model effectively suppresses this prior bias, correctly reasoning that the events occur \textit{"At the same time"} (Option C). 
This highlights our method's ability to correct temporal dependency errors where the model defaults to a linear narrative.}
\label{fig:surgery example2}
\end{figure}

In this section, we present qualitative comparisons to intuitively demonstrate the effectiveness of our Temporal Head Amplification.
As shown in \figref{fig:surgery example1} and \figref{fig:surgery example2}, we display the key video frames, the input question, and the model's responses under two conditions: (1) the \textit{Baseline} (without intervention), and (2) our \textit{Intervened} model (with temporal head amplification).
These cases highlight how our circuit-guided intervention effectively corrects temporal hallucinations and rectifies errors in chronological reasoning, enabling the model to accurately capture action sequences and event dependencies where the baseline fails.

\begin{minipage}[!t]{\linewidth}
\begin{minipage}[!t]{0.5\linewidth}
\centering
\renewcommand{\arraystretch}{1.1}
\scriptsize
\tabcolsep=1.00mm
\begin{tabular}{lcc}
    \hline
    Head Type & Frames=16 & Frames=32 \\ \hline
    Baseline & 73.6 & 74.6 \\
    (mass, neff))& 73.6 & 74.2 \\ 
    (mass, $\sigma$)  & 73.7& 74.8 \\
    (neff, $\sigma$) & 73.4 & 74.1 \\ 
    Random & 73.2 & 74.2 \\ 
    Temporal head & 75.2 & 75.5 \\ \hline
\end{tabular}
\makeatletter\def\@captype{table}\makeatother\caption{Ablation study of temporal head type.}
\label{tab:temporal_head_layers}
\end{minipage}
\begin{minipage}[!t]{0.45\linewidth}
\centering
\includegraphics[width=.75\linewidth]{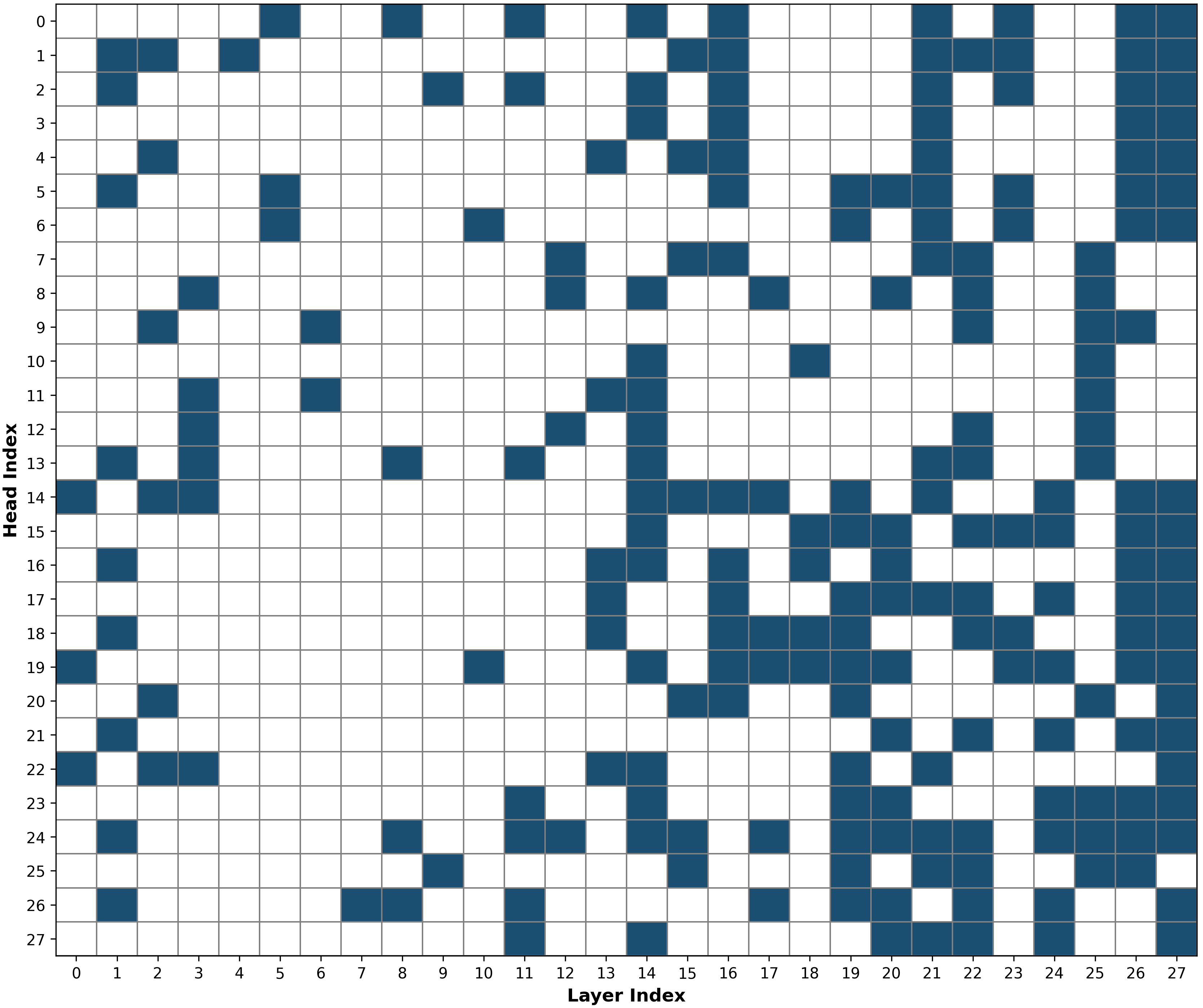}
\makeatletter\def\@captype{figure}\makeatother\caption{Layer head distribution heatmap.}\label{fig:head_heatmap}
\end{minipage}
\end{minipage}

\end{document}